%% file: paper.tex
\pdfoutput=1

\documentclass[10pt,twocolumn,letterpaper]{article}

\usepackage{iccv}
\usepackage{times}
\usepackage{epsfig}
\usepackage{xcolor}
\usepackage{placeins}
\usepackage{float}
\usepackage{multirow}
\usepackage{graphicx}
\usepackage{amsmath}
\usepackage{mathtools}
\usepackage{amssymb}
\usepackage{subcaption}
\usepackage{tabulary}
\usepackage{booktabs} 
\usepackage{natbib}
\usepackage[final]{microtype}
\usepackage[toc,page]{appendix}
\usepackage{siunitx}
\usepackage[accsupp]{axessibility}
\setcitestyle{numbers}
\setcitestyle{square}

\usepackage[breaklinks=true,bookmarks=false,hyperfootnotes=false]{hyperref}

\iccvfinalcopy % *** Uncomment this line for the final submission

\def\iccvPaperID{10711} % *** Enter the ICCV Paper ID here

\newcommand\blfootnote[1]{%
  \begingroup
  \renewcommand\thefootnote{}\footnote{#1}%
  \addtocounter{footnote}{-1}%
  \endgroup
}

\begin{document}

%%%%%%%%% TITLE
\title{e-ViL: A Dataset and Benchmark for Natural Language Explanations in Vision-Language Tasks}

\author{
	\textbf{Maxime Kayser}\textsuperscript{1*} \ \ \ 
	\textbf{Oana-Maria Camburu}\textsuperscript{1} \ \ \ 
	\textbf{Leonard Salewski}\textsuperscript{2} \ \ \ 
	\textbf{Cornelius Emde}\textsuperscript{1}\\
	\textbf{Virginie Do}\textsuperscript{1**} \ \ \
	\textbf{Zeynep Akata}\textsuperscript{2,3,4} \ \ \
	\textbf{Thomas Lukasiewicz}\textsuperscript{1} \\\\
	\textsuperscript{1}Department of Computer Science, University of Oxford \ \ \ \ \ \textsuperscript{2}University of Tübingen \\ \textsuperscript{3}Max Planck Institute for Intelligent Systems  \ \ \ \ \ \textsuperscript{4}Max Planck Institute for Informatics %\ \ \ \ \ \textsuperscript{3}Université Paris Dauphine \ \ \ \ \ \textsuperscript{4}Facebook AI Research
}

\maketitle
\blfootnote{*Corresponding Author: maxime.kayser@cs.ox.ac.uk}
\blfootnote{**Now at Université Paris-Dauphine, PSL, and Facebook AI Research.}
\ificcvfinal\thispagestyle{empty}\fi

%%%%%%%%% ABSTRACT
\begin{abstract}
    \input{sections/abstract}
\end{abstract}

%%%%%%%%% BODY TEXT
\vspace{-2ex}
\section{Introduction} \label{sec:intro}
    \input{sections/intro.tex}

\section{Related Work} \label{sec:rw}
    \input{sections/relwork.tex}

\section{The e-SNLI-VE Dataset} \label{sec:esnlive}
    \input{sections/esnlive.tex}

\section{The e-ViL Benchmark} \label{sec:evil}
    \input{sections/benchmark}

\section{Experimental Evaluation} \label{sec:exp}
    \input{sections/experiments}

\section{Summary and Outlook}
    \input{sections/conclusion}
    
\section*{Acknowledgements}
    \input{sections/acknowledge}
%-------------------------------------------------------------------------

%------------------------------------------------------------------------
% \section{Final copy}

% You must include your signed IEEE copyright release form when you submit
% your finished paper. We MUST have this form before your paper can be
% published in the proceedings.

{\small
\bibliographystyle{plainnat} %ieee_fullname}
\bibliography{references, egbib}
}

\clearpage
\begin{appendices}
    \section{Qualitative Examples} \label{app:ex}
        \input{appendix/examples}
    \section{e-SNLI-VE} \label{app:dset}
        \input{appendix/eSNLIVE}
    \section{Benchmark Models} \label{app:models}
        \input{appendix/models}
    \section{Human Evaluation Framework} \label{app:mturk}
        \input{appendix/eval}
    \section{Results} \label{app:results}
        \input{appendix/results}

\end{appendices}

\end{document}

%% file: sections/abstract.tex
Recently, there has been an increasing number of efforts to introduce models capable of generating natural language explanations (NLEs) for their predictions on vision-language (VL) tasks. Such models are appealing, because they can provide human-friendly and comprehensive explanations. However, there is a lack of comparison between existing methods, which is due to a lack of re-usable evaluation frameworks and a scarcity of datasets. In this work, we introduce e-ViL and e-SNLI-VE. e-ViL is a benchmark for explainable vision-language tasks that establishes a unified evaluation framework and provides the first comprehensive comparison of existing approaches that generate NLEs for VL tasks. It spans four models and three datasets and both automatic metrics and human evaluation are used to assess model-generated explanations. e-SNLI-VE is currently the largest existing VL dataset with NLEs (over 430k instances). We also propose a new model that combines UNITER \cite{chen_uniter_2020}, which learns joint embeddings of images and text, and GPT-2 \cite{radford2019language}, a pre-trained language model that is well-suited for text generation. It surpasses the previous state of the art by a large margin across all datasets. Code and data are available here: \url{https://github.com/maximek3/e-ViL}.

%% file: sections/intro.tex
Deep learning models achieve promising performance across a variety of tasks but are typically black box in nature. There are several arguments for making these models more explainable. For example, explanations are crucial in establishing trust and accountability, which is especially relevant in safety-critical applications such as healthcare or autonomous vehicles. They can also enable us to better understand and correct the learned biases of models \cite{arrieta2020explainable}. Explainability efforts in vision tasks largely focus on highlighting relevant regions in the image, which can be achieved via tools such as saliency maps \cite{ancona2018towards} or attention maps \cite{xiao2015application}. Our work focuses on natural language explanations (NLEs), which aim to explain the decision-making process of a model via generated sentences. Besides being easy to understand for lay users, NLEs can explain more complex and fine-grained reasoning, which goes beyond highlighting the important image regions. We compare different models that generate NLEs for vision-language (VL) tasks, i.e., tasks where the input consists of visual and textual information, such as visual question-answering (VQA). 

NLEs for VL tasks (VL-NLE) is an emerging field, and only few datasets exist. Moreover, existing datasets tend to be relatively small and unchallenging (e.g., VQA-X~\cite{park_multimodal_2018}) or noisy (e.g., VQA-E~\cite{li_vqa-e_2018}). Another limitation of the VL-NLE field is that there is currently no unified evaluation framework, i.e., there is no consensus on how to evaluate NLEs. NLEs are difficult to evaluate, as correct explanations can differ both in syntactic form and in semantic meaning. For example, ``Because she has a big smile on her face” and ``Because her team just scored a goal” can both be correct explanations for the answer ``Yes” to the question ``Is the girl happy?”, but existing automatic natural language generation (NLG) metrics are poor at capturing this. As such, the gold standard for assessing NLEs is human evaluation. Past work have all used different approaches for human evaluations, and therefore no objective comparison exists. %This makes it difficult to know which architectures are most effective. % and worth developing in the future.

In this work, we propose five main contributions to address the lack of comparison between existing work. (1)~We propose e-ViL, the first benchmark for VL-NLE tasks. e-ViL spans across three datasets of human-written NLEs, and provides a unified evaluation framework that is designed to be re-usable for future works. (2) Using e-ViL, we compare four VL-NLE models. (3) We introduce e-SNLI-VE, a dataset of over 430k instances, the currently largest dataset for VL-NLE. (4) We introduce a novel model, called e-UG, which surpasses the state of the art by a large (and significant) margin across all three datasets. (5) We provide the currently largest study on the correlation between automatic NLG metrics and human evaluation of NLEs.

% \begin{enumerate}
%     \vspace{-1ex}
%     \item We propose e-ViL, the first benchmark for VL-NLE tasks. e-ViL spans across three datasets of human-written NLEs, and provides a unified evaluation framework that is designed to be re-usable for future works. 
%     \vspace{-1ex}
%     \item Using e-ViL, we compare four VL-NLE models.
%     \vspace{-1ex}
%     \item We introduce e-SNLI-VE, a dataset of over 430k instances, the currently largest dataset for VL-NLE. 
%     \vspace{-1ex}
%     \item We introduce a novel model, called e-UG, which surpasses the state-of-the-art by a large (and significant) margin across all three datasets.
%     \vspace{-1ex}
%     \item We provide the currently largest study on the correlation between automatic NLG metrics and human evaluation of NLEs. 
% \end{enumerate}

%
%\vspace{-1ex}
%\begin{enumerate}
%    \item  We propose e-ViL, the first benchmark for VL-NLE tasks. e-ViL spans across three datasets of human-written NLEs, and provides a unified evaluation %framework that is designed to be re-usable in future works.
%    
%    \vspace{-1ex}
%    \item Using e-ViL, we compare four VL-NLE models.
%    
%    \vspace{-1ex}
%    \item We introduce e-SNLI-VE, a dataset of over 430k instances, currently the largest VL-NLE dataset.
%    
%    \vspace{-1ex}
%    \item We introduce a novel model, called e-UG, which surpasses the state of the art by a large margin across all three datasets.
%\end{enumerate}

%% file: sections/relwork.tex
\begin{figure}
\begin{center}
\includegraphics[width=0.8\linewidth]{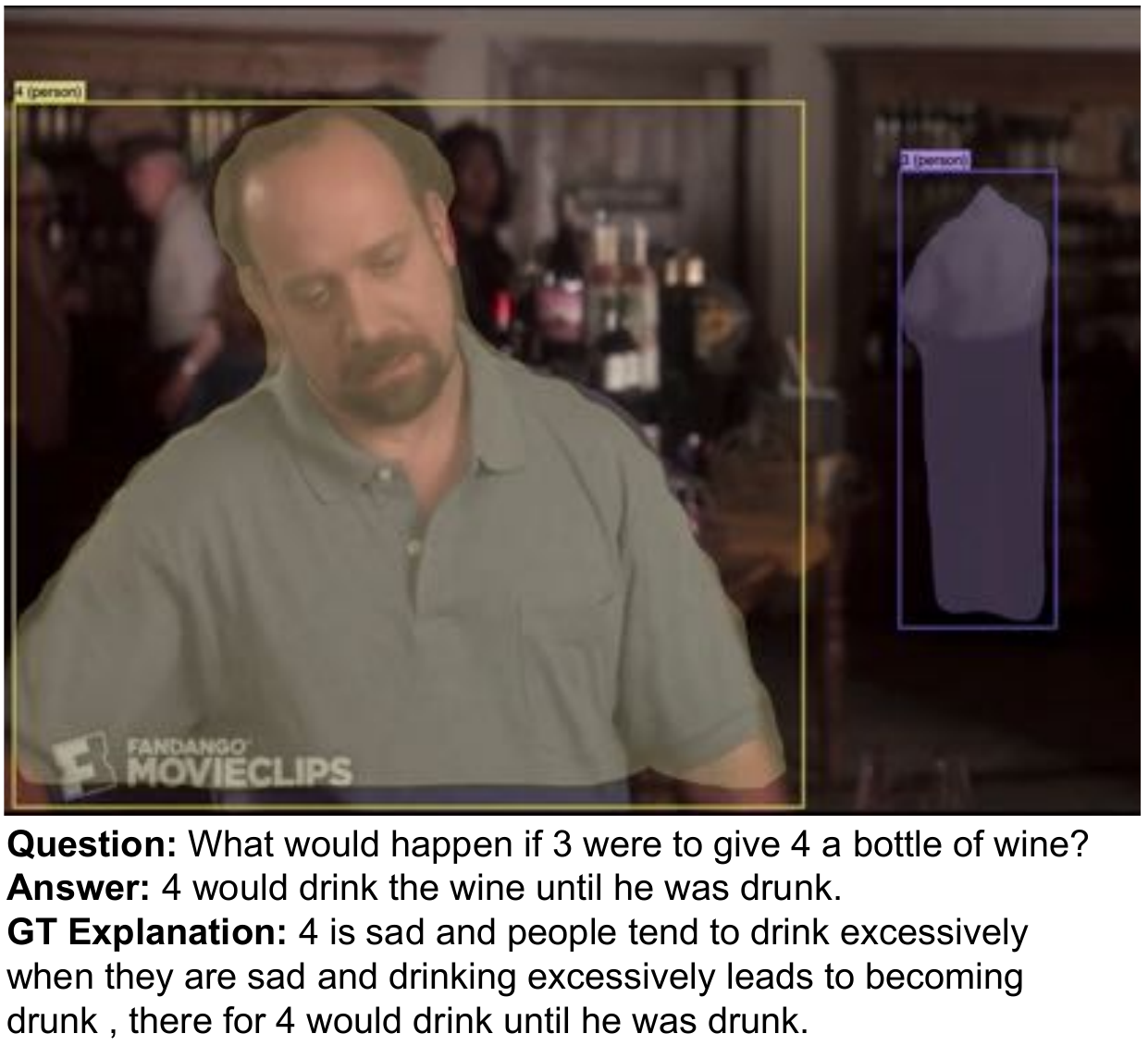}
\end{center}\vspace{-1ex}
    \caption{VCR images require commonsense reasoning that often goes beyond the visual content of the image.}%
    \label{fig:vcr_comsen}
\vspace{-2ex}
\end{figure}

\paragraph{Explainability in Computer Vision.}
Common approa\-ches to explainability of deep learning methods in computer vision are saliency maps~\cite{ancona2018towards}, attention maps~\cite{xiao2015application}, and activation vectors~\cite{kim_interpretability_2018}. Saliency and attention maps indicate where a model looks. This may tell us what regions of the image are most important in the decision-making process of a model. Activation vectors are a way to make sense of the inner representation of a model, e.g., by mapping it to human-known concepts. However, these approaches often cover only a fraction of the reasoning of a model. On the contrary, NLEs can convey higher-order reasoning and describe complex concepts. For example, in Figure \ref{fig:vcr_comsen}, highlighted image regions or weights for different concepts would not be sufficient to explain the answer. Additionally, it has been shown that numerical or visual explanatory methods, in some cases, can be confusing even for data scientists~\cite{kaur} and can pose problems even for explaining trivial models \cite{DBLP:journals/corr/abs-2009-11023}. %For example, explaining the decision of a self-driving car to accelerate by the NLE ``I accelerated because we are on a highway with light traffic, there are good weather conditions, and I am at speed 50mph, which is still below the speed limit of 70mph for this road.'' is more straightforward than a combination of heatmaps over the whole environment, especially when parts of the decision are not present in the environment at a given moment (e.g., the speed limit sign).

\vspace*{-3ex}
\paragraph{NLEs.}
First adoptions of NLEs have been in image classification~\cite{hendricks2016generating} and were further extended to self-driving cars \citep{kim2018textual}, VQA \cite{park_multimodal_2018}, and natural language processing \citep{camburu_e-snli_2018, explainyourself, bhagavatula2020abductive, atanasova-etal-2020-generating-fact, kotonya-toni-2020-explainable, narang_wt5_2020, kumar_nile_2020, camburu-etal-2020-make}. %Prominent works on VL-NLE was done by~\citet{park_multimodal_2018}, \citet{wu_faithful_2019}, \citet{}.
The most important works in VL-NLE \citep{park_multimodal_2018, wu_faithful_2019, marasovic_natural_2020} are included in this benchmark.
%Their Pointing and Justification Explanation (PJ-X) model generates a multimodal representation of the input image and text, which is used to predict the answer and produce a visual attention map over the image. An LSTM-based decoder~\cite{hochreiter1997long} then generates an explanation from the multimodal representation, attention map (\alert{xxVerify}), and answer. In \citet{li_vqa-e_2018} the answer and explanation are generated simultaneously. The model FME by~\citet{wu_faithful_2019} uses the UpDown architecture~\cite{anderson_bottom-up_2018} as a backbone for obtaining the multimodal representation of the input, which, together with the answer, is fed to a double LSTM explanation generator. \citet{marasovic_natural_2020}  use either an object detector, a situation recognition model, or VisualComet \alert{xxCite} as backbone, and then generate the explanation using GPT-2, a pretrained language model \cite{radford2019language}. However, they do not predict the answers for task itself, but condition the explanations on the ground-truth answer. Their visual representation of the image is not trained, but fixed. \citet{ayyubi_generating_2020} tackle explanation generation with ViLBERT~\cite{lu_vilbert_2019} and GPT-2. %\alert{xxOurOwnModel}

\vspace*{-3ex}
\paragraph{VL-NLE Datasets.}
Existing models learn to generate NLEs in a supervised manner and, therefore, require training sets of human-written explanations. Besides the image classification datasets ACT-X \cite{park_multimodal_2018} and CUB \cite{wah2011caltech, hendricks2016generating}, and the video dataset BDD-X \cite{kim2018textual}, there are currently three VL datasets with NLEs. % on top of VL tasks.
The VQA-X dataset \cite{park_multimodal_2018} was introduced first and provides NLEs for a small subset of questions from VQA v2 \cite{VQA}. It consists of 33k QA pairs (28k images). However, many of the NLEs in VQA-X are trivial and could be guessed without looking at the image. For example, ``because she is riding a wave on a surfboard'' is an NLE for the answer ``surfing'' to the question ``What is the woman in the image doing?'' that can easily be guessed from the answer, without looking at the image (more examples are given in Figure \ref{fig:dset_examples}). 
VQA-E~\cite{li_vqa-e_2018} is another dataset that also builds on top of VQA v2. However, its explanations are gathered in an automatic way and were found to be of low quality by \citet{marasovic_natural_2020}, where the model-generated explanations obtain a human evaluation accuracy\footnote{Percentage of explanations that, given the image and question, support the predicted answer.} that is only 3\% short of the VQA-E ground-truth explanations (66.5\%), suggesting that the dataset is essentially solved. It is therefore not used in our benchmark. Finally, the VCR dataset \cite{zellers_recognition_2019} provides NLEs for VQA instances that require substantial commonsense knowledge (see Figure \ref{fig:vcr_comsen}). The questions are challenging, and therefore both the answers and NLEs are given in the form of multiple-choice options. %In the questions and answers of VCR, people and objects are linked to specific bounding boxes in the image. 
%VCR also differs from other VL datasets in that it provides direct links between the people and objects in the question and the image. 

Our proposed dataset, e-SNLI-VE, extends the range of VL-NLE datasets and addresses some of the prior limitations. It contains over 430k instances for which the explanations rely on the image content (see examples in Figure \ref{fig:dset_examples}). We will describe the dataset in more detail in Section \ref{sec:esnlive}.

\vspace*{-3.5ex}
\paragraph{Evaluations and Comparisons.}
Evaluating NLG is challenging and a much-studied field~\cite{gatt2018survey}. Evaluating NLEs is even more difficult, as sentences may not only differ in their syntactic form but also in their semantic meaning, e.g., there can be several different reasons why a sentence contradicts an image. For this reason, current automatic NLG metrics, such as the BLEU score~\cite{papineni2002bleu}, do not perform well in evaluating NLEs~\cite{camburu_e-snli_2018}. Hence, several works have used human evaluation to assess their generated explanations \cite{park_multimodal_2018,wu_faithful_2019,marasovic_natural_2020, dua_beyond_2020}. However, they all used different evaluation rules, preventing one from being able to compare existing VL-NLE models. The main differences lie in the datasets used, the questions asked to annotators, whether the assessment is absolute or based on a ranking, and the formula used to calculate the final score. We select the best practices from existing evaluation schemes and develop a unified and re-usable human evaluation framework for VL-NLE.

\vspace{-1.5ex}

%% file: sections/esnlive.tex
\vspace{-1ex}
We introduce e-SNLI-VE, a large-scale dataset for visual-textual entailment with NLEs. We built it by merging the explanations from e-SNLI \cite{camburu_e-snli_2018} and the image-sentence pairs from SNLI-VE \cite{xie_visual_2019}. We use several filters and manual relabeling steps to address the challenges that arise from merging these datasets. The validation and test sets were relabelled by hand. The dataset is publicly available\footnote{\url{https://github.com/maximek3/e-ViL}}.

\begin{table*}[htbp]
    \begin{center}
    \begin{tabulary}{\linewidth}{LCCC}
    \toprule
                                     & Train          & Validation     & Test           \\
    \midrule
    \# Image-Hypothesis pairs (\# Images)       & 401,717 (29,783)       & 14,339 (1,000)         & 14,740 (1,000)        \\
    %\# Images                        & 29,783          & 1,000           & 1,000           \\
    Label distribution (C/N/E, \%)   & 36.0 / 31.3 / 32.6 & 39.4 / 24.0 / 36.6 & 38.8 / 25.8 / 35.4 \\
    Mean hypothesis length (median)  & 7.4 (7)       & 7.3 (7)        & 7.4 (7)       \\
    Mean explanation length (median) & 12.4 (11)     & 13.3 (12)     & 13.3 (12)     \\
    \bottomrule
    \end{tabulary}
  \caption{e-SNLI-VE summary statistics. C, N, and E stand for Contradiction, Neutral, and Entailment, respectively.}%
  \label{tab:esnlive}
  \end{center}
  \vspace{-4ex}
\end{table*}

%NOTE: should we also release a new SNLI-VE, which contains only the processing steps that dealt with labels and not explanations? Their dataset has been cited/used 34 times.

\vspace*{-0.5ex}
\subsection{Correcting SNLI-VE} \label{d:fn}
\vspace*{-0.5ex}

\begin{figure}[t]
\begin{center}
\includegraphics[width=0.8\linewidth]{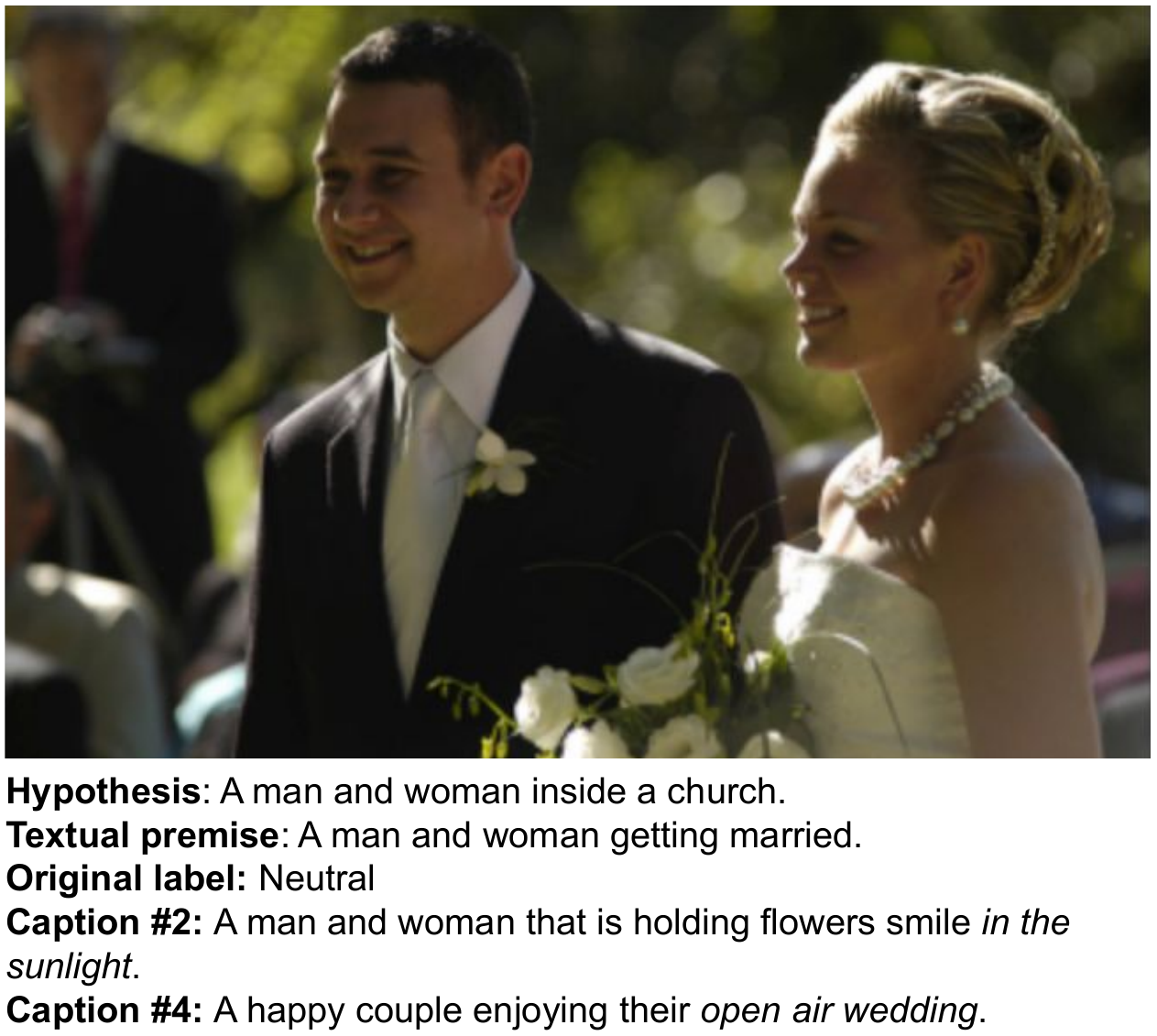}
\end{center}\vspace{-1ex}
   \caption{The original label of the textual premise-hypothesis pair in SNLI is neutral. However, by considering alternative captions describing the same image (\#2 and \#4), we can deduct that the neutral label is false.}
\label{fig:fn}
\vspace{-2ex}
\end{figure}

In SNLI-VE \cite{xie_visual_2019}, an image and a textual hypothesis are given, and the task is to classify the relation between the image-premise and the textual hypothesis. The possible labels are \emph{entailment} (if the hypothesis is true, given the image), \emph{contradiction} (the hypothesis is false, given the image), or \emph{neutral} (if there is not enough evidence to conclude whether the hypothesis is true or false). SNLI-VE builds off the SNLI \cite{bowman2015large} dataset, by replacing textual premises with Flickr30k images \cite{young2014image}. This is possible, because the textual premises in SNLI are caption sentences of those images. However, this replacement led to labeling errors, as an image typically contains more information than a single caption describing it. Especially for the neutral class, a caption may not have enough evidence to suggest entailment or contradiction, but the corresponding image does (see Figure \ref{fig:fn}). On a manually evaluated subset of 535 samples, we found a 38.6\% error rate among the neutral labels. This subset will be used below to evaluate the effectiveness of our filters. Error rates for entailment and contradiction are reported to be under~1\%~\cite{xie_visual_2019}, hence we focus only on correcting the neutral instances.

In the validation and test sets, we relabeled the neutral examples using Amazon Mechanical Turk (MTurk). To ensure high-quality annotations, we used a series of quality control measures, such as in-browser checks, inserting trusted examples, and collecting three annotations per instance. In total, 39\% of the neutral labels were changed to entailment or contradiction. The label distribution shifted from uniform to Ent/Neut/Cont of 39\%/20\%/41\% and 39\%/21\%/40\% for the validation and test sets, respectively.

For the training set, we propose an automatic way to remove false neutrals. We discovered that the five captions that come with each image often provide clues whether the label is indeed neutral, or not. For every image-hypothesis pair $i$, we ran a natural language inference model $m_{\mathrm{nli}}$ on each caption-hypothesis pair $p_{i,c}$, where $c$ is one of the captions. If the original label of image-hypothesis pair $i$ is neutral, but  $\sum_{c} m_{\mathrm{nli}}(p_{i,c})$ indicates with high confidence that the label is not neutral, we deem the label incorrect and removed the instance from the dataset. An example is shown in Figure \ref{fig:fn}. For $m_{\mathrm{nli}}$, we used Roberta-large \cite{liu2019roberta} trained on the MNLI dataset \cite{williams2018broad}. Instances were removed if $\sum_{c} m_{\mathrm{nli}}(p_{i,c})$ exceeded $2.0$ for entailment and contradiction classes. On our 535-samples subset, this filter decreased the error of neutral labels from 39\% to 24\%. When validated against the relabeling on the validation set, the error decreased from~39\%~to~30\%.

\vspace*{-0.5ex}
\subsection{Adding Explanations to SNLI-VE}
\vspace*{-0.5ex}

To create e-SNLI-VE, we source explanations from e-SNLI \cite{camburu_e-snli_2018}, which extends SNLI with human-written NLEs. However, the explanations in e-SNLI are tailored to the textual premise-hypothesis pairs and are therefore not always well-suited for the image-hypothesis pair. After simply merging both datasets, we found that initially 36\%, 22\%, and 42\% of explanations were of low (incorrect), medium (correct, but there is an obvious better choice), and high quality (correct and relevant), respectively. We propose several steps to detect and remove explanations of low and medium quality. The filters were designed to ensure an optimal trade-off between precision and recall (for flagging bad explanations) and with the constraint that the final dataset remains reasonably balanced.

\vspace{-3ex} 
\paragraph{Re-annotation.} \label{d:reanno}
First, we replace the explanations for the neutral pairs in the validation and test sets with new ones, collected via MTurk at the same time as we collected new labels for these subsets. In order to submit the annotation of an image-sentence pair, workers must choose a label, highlight words in the hypothesis, and use at least half of the highlighted words in the explanation. %On a random sample of 100 explanations, we found that 83\% of the newly annotated explanations are of high quality. %(?seems like a low number, should we still mention it?). Details of the MTurk re-annotation are given in Appendix \ref{app:mturk_relab}.

\vspace{-3ex}
\paragraph{Keyword Filter.} \label{d:kw}
Next, we use keyword filtering to detect explanations that make reference to a linguistic feature of the textual premise. The keywords, which we manually defined, are ``synonym", ``mention", ``rephrasing", ``sentence", ``way to say", and ``another word for". The keyword filter removed 4.6\% of all instances, and our 535-samples subset suggests that \emph{all} filtered explanations were indeed of low quality.

\vspace{-3ex}
\paragraph{Similarity Filter.} \label{d:sim}
We noticed that the share of low-quality explanations is highest for entailment examples. This happens frequently when the textual premise and hypothesis are almost identical, as then the explanation often just repeats both statements. To overcome this, we removed all examples where the ROUGE-1 score (a measure for sentence similarity \cite{lin2004rouge}) between the textual premise and hypothesis was above 0.57. This reduced the share of low-quality explanations for entailment by 4.2\%.

\vspace{-3ex}
\paragraph{Uncertainty Filter.} \label{d:unc}
Lastly, we found that image-hypo\-the\-sis pairs with high uncertainty are correlated with low-quality explanations for contradictions. We define uncertainty as the diversion of the scores from $m_{\mathrm{nli}}(p_{i,c})$ for the five image captions. $m_{\mathrm{nli}}$ is the same Roberta-large model that was described above. This filter reduced the share of low-quality explanations for contradiction examples by 5.1\%. 

The final e-SNLI-VE dataset statistics are displayed in Table~\ref{tab:esnlive}. An additional evaluation of e-SNLI-VE by external annotators, and a comparison with existing VL-NLE datasets, is provided in Table \ref{tab:dataQual}. The results indicate that the quality of the e-SNLI-VE ground-truth explanations is not far off the human-annotated VQA-X and VCR datasets. Qualitative examples and a more detailed rundown of our filtering methods are in Appendix \ref{app:dset}.

%% file: sections/benchmark.tex
In this section, we introduce the VL-NLE task, describe how explanations are evaluated in e-ViL, and describe the datasets covered in our benchmark. 

\subsection{Task Formulation}

We denote a module that solves a VL task as $M_T$, which takes as input visual information $V$ and textual information~$L$. 
Its objective is to complete a task $T$ where the outcome is $a$, i.e., $M_T(V,L)=a$. An example of a VL task is VQA, where $V$ is an image, $L$ is a question, and $T$ is the task of providing the answer~$a$ to that question. We extend this by an additional task $E$, which requires an NLE $e$ justifying how $V$ and $L$ lead to~$a$, solved by the module $M_E(V,L)=e$. The final model $M$ then consists of $M_T$ and $M_E$. Thus, $M = (M_T, M_E)$ and $M(V, L)=a,e$. 

\subsection{Datasets}
Our benchmark uses the following three datasets, which vary in size and domain. Examples are shown in Figure \ref{fig:dset_examples} in the appendix.

\vspace{-3ex}
\paragraph{e-SNLI-VE.}
Our proposed e-SNLI-VE dataset %consists of over 430k image-hypothesis pairs, where the task $T$ is to classify the pairs as entailment, neutral, or contradiction. More details on the dataset can be found
has been described in Section~\ref{sec:esnlive}.

\vspace{-3ex}
\paragraph{VQA-X.}
VQA-X \cite{park_multimodal_2018} contains human written explanations for a subset of questions from the VQA v2 dataset \cite{goyal2017making}. The image-question pairs are split into train, dev, and test with 29.5k, 1.5k, and 2k instances, respectively. %VQA-X only includes questions that require at least the intellect of an average 9 year old. This makes sure that overly easy questions, such as ``What color is the car?", are excluded. 
The task $T$ is formulated as a multi-label classification task of 3,129 different classes. One question can have multiple possible answers.% as each example has been annotated by multiple people.

\vspace{-3ex}
\paragraph{VCR.}
Visual Commonsense Reasoning (VCR) is a VL dataset that asks multiple-choice (single answer) questions about images from movies~ \cite{zellers_recognition_2019}. In addition to four answer options, it also provides four NLEs options, out of which one is correct. For the purpose of our proposed VL-NLE task, we reformulate it as an explanation generation task. As the test set for VCR is not publicly available, we split the original train set into a train and dev set, and use the original validation set as test set. The splits are of size  191.6k, 21.3k, and 26.5k, respectively.

\vspace{-3ex}
\paragraph{Human Evaluation of Datasets.}
In our benchmark experiments (Section~\ref{sec:exp}), human annotators evaluate the ground-truth explanations of all three datasets. For each dataset, 300 examples are evaluated by 12 annotators each, resulting in 3,600 evaluations. %For each explanation, participants responded to the question ``Given the image and the question/hypothesis, does the explanation justify the answer?" with \textit{no}, \textit{weak no}, \textit{weak yes} or \textit{yes}. 
The results in Table~\ref{tab:dataQual} show that e-SNLI-VE comes close to the manually annotated datasets VCR and VQA-X (82.8\% explanations with \textit{yes} or \textit{weak yes} vs.\ 87.9\% and 91.4\%). Besides the use of effective, but imperfect, automatic filters, another explanation for the higher share of noise is the trickiness (out of a 100 human re-annotated explanations for neutral examples, we found that 17\% had flaws, identical to the share of (weak) no in Table~\ref{tab:dataQual}) and ambiguity (when three of us chose the labels for a set of 100 image-hypothesis pairs, we only had full agreement on 54\% of examples) inherent in the e-SNLI-VE task.

\begin{table}[htbp] 
    \begin{center}
    \begin{tabulary}{\linewidth}{LCCCC}
    \toprule
              & No   & Weak No & Weak Yes & Yes  \\
    \midrule
    e-SNLI-VE & 10.3\% & 6.9\%   & 27.7\%   & 55.1\% \\
    VQA-X     & 4.1\%  & 4.5\%   & 25.1\%   & 66.3\% \\
    VCR       & 6.9\%  & 5.2\%   & 36.6\%   & 51.3\% \\
    \bottomrule
    \end{tabulary}
  \caption{Human evaluation of the ground-truth explanations for the three datasets used in e-ViL. The question asked was: ``Given the image and the question/hypothesis, does the explanation justify the answer?".
  }%
  \label{tab:dataQual}
  \end{center}
  \vspace{-6ex}
\end{table}

\subsection{Evaluation} \label{sec:tf}
\paragraph{Evaluation Scores.}
We define separate evaluation scores $S_T$, $S_E$, and $S_O$ for $M_T$, $M_E$, and $M$, respectively. $S_T$ is the metric that is defined by the original VL task $T$, e.g., label accuracy for e-SNLI-VE and VCR, and VQA accuracy for VQA-X. 
We define $S_E$ as the average explanation score of the examples for which the answer $a$ was predicted correctly. In line with previous work~\cite{park_multimodal_2018, wu_faithful_2019, marasovic_natural_2020}, we for now assume the simplified scenario that an explanation is always false when it justifies an incorrect answer. The explanation score can be any custom human or automatic metric. Due to the limitations of current automated NLG metrics for evaluating NLEs, we developed a human evaluation framework for computing $S_E$, outlined in the paragraph below.
%An explanation $e$ is expected to be false when the answer $a$ is predicted incorrectly (as it is expected to justify a wrong answer) and is, therefore, not considered in the computation of $S_E$. 
Finally, we want $S_O$ to summarize the performance of a model on both tasks $T$ and $E$, to give us the overall performance of a VL-NLE model $M$. We define $S_O = S_TS_E$, which equates to the average of the scores of all explanations, but where we set the score of an explanation to~$0$ if its associated answer was predicted incorrectly. This can also be viewed the explanation score $S_E$ multiplied by a coefficient for task performance (accuracy, in most cases). We introduced  this measure to avoid giving an advantage to models that purely optimize for generating a few good explanations but neglect the task itself.
% We compute $S_O$ by getting the average of all explanations, but setting the score of an explanation to $0$ if the answer was predicted incorrectly. 
% This equates to $S_O = S_E \times C/(C+I)$, where $C$ and $I$ are the number of samples where the answer was correct and incorrect, respectively. %For VQA-X, we only evaluate explanations when the model predicted the answer with the highest score, as ground-truth explanations are only given for that answer. 
% For VCR and e-SNLI-VE, where $S_T$ is the accuracy, we get $S_O= S_T \times S_E$.

%As mentioned before, existing automated NLG metrics have strong limitations for evaluating NLEs. Therefore, to compute $S_E$, we developed the human evaluation framework outlined below.

\vspace{-3ex}
\paragraph{Human Evaluation Framework.} 
We collect human annotations on MTurk, where we ask the annotators to proceed in two steps. First, they have to solve the task $T$, i.e., provide the answer $a$ to the question. This ensures the annotators reflect on the question first and enables us to do in-browser quality checks (since we know the answers). We disregard their annotation if they answered the VL task $T$ incorrectly.
%However, since we cannot assume that all answers in the datasets are correct, we only enforce that \alert{XX} out of four instances that an annotator had to %solve in one submission (a HIT) should coincide with the answer that we have in our datasets. 
%After the annotator provided the answer, we present her with two explanations: one generated by a model and the ground truth from our dataset. The annotators do not know which explanation is the ground truth or the generated one. We provide the ground-truth explanations for comparison, as well as for implicitly anchoring the annotators to a priori high-quality explanations. 

For each explanation, we ask them a simple evaluation question: ``Given the image and the question/hypothesis, does the explanation justify the answer?". %We argue that the main purpose of NLEs is to be a form of model explanation that lay users can easily understand and agree with and we hope that this simple question captures this. 
We follow \citet{marasovic_natural_2020} in giving four response choices: \emph{yes}, \emph{weak yes}, \emph{weak no}, and \emph{no}. 
%Our experiences have also shown that requiring annotators to decide between leaning towards \textit{yes} or \textit{no} is important, as otherwise they too often take a shortcut and select the neutral option. 
%\citet{marasovic_natural_2020} later merges \emph{weak yes} with \emph{yes} and \emph{weak no} with \emph{no}, giving them only a binary evaluation for every explanation. We argue that keeping the four response options will allow for a more fine-grained evaluation. 
We map \emph{yes}, \emph{weak yes}, \emph{weak no}, and \emph{no} to the numeric scores of $1$, $\frac{2}{3}$, $\frac{1}{3}$, and $0$, respectively. %This enables us to get an overall score that captures more nuanced performance differences between the models. 

We also ask annotators to select the main shortcomings (if any) of the explanations. We observe three main limitations of explanations. First, they can \textit{insufficiently justify the answer}. For example, the sentence ``because it's cloudy" does not sufficiently justify the answer ``the sea is not calm". Second, an explanation can \textit{incorrectly describe the image}, e.g., if a model learned generic explanations that are not anchored in the image. ``There is a person riding a surfboard on a wave" is generally a good explanation for the answer ``surfing" when asked ``what activity is this?", but the image may actually display a dog surfing. Lastly, the sentences can be \textit{nonsensical}, such as ``a man cannot be a man". 

For each model-dataset pair, we select a random sample of 300 datapoints where the model answered the question correctly. %We do not evaluate explanations where the model answered incorrectly, as a justification of a wrong answer is not expected to be correct. 
Every sample contains only unique images. For VCR, all movies are represented in the samples. Note that it is not possible to evaluate all models on exactly the same instances, as they do not all answer the same questions correctly. Taking a subset of examples where \emph{all} models answered correctly is disadvantageous for two reasons. First, this makes the benchmark less re-usable, as future methods might not answer the same questions correctly. Second, this would bias the dataset towards the questions that the weakest model answered correctly. However, in order to still maximize the overlap between the samples, we shuffled all the instances in the test sets randomly and then for each model we took the 300 first on which the answer was correct. %Still, we designed a procedure that maximizes the overlap between the different sets, without models influencing each other. %do we need to lay this out or not mention at all?. Our overlaps are xxx.

We propose several measures to further ensure robustness and re-usability of the framework. In order to account for annotator subjectivity, we evaluate every instance by three different annotators. The final score per explanation is given by the average of all evaluations. In addition, we evaluate one model at a time to avoid potential anchoring effects between models (e.g., the annotator evaluates one model more favorably,  because they are influenced by poor explanations from a different model). %\footnote{We cannot enforce that the same annotator completes tasks for more than one model}. 
To implicitly induce a uniform anchoring effect, the annotators evaluate both the ground-truth explanation (which is invariant to the model) and the explanation generated by a model for every image-question pair. They do not know which is which and are not asked to compare them. This implicitly ensures that all evaluations have the same anchor (the ground-truth) and it allows us to compute $S_E$ in different ways, as outlined in Appendix \ref{sec:evil_alts}. %This also enables us to detect workers that did not understand the task well enough. 
Overall, over 200 individual annotators were employed for the benchmark and all of them had to have a 98\% prior acceptance rate on MTurk. Finally, we bolster our results with statistical tests in Appendix~\ref{sec:stat_analysis}.

More details and screenshots of our MTurk evaluation can be found in  Appendix \ref{app:results}. For re-usability, we publicly release the questionnaires used in our benchmark\footnote{\url{https://github.com/maximek3/e-ViL}}.

% NOTE: This belongs to the next section, but it needs to be placed here such that it appears on page 6
\begin{figure*}[ht] 
    \begin{subfigure}[h]{0.4\linewidth}
        \centering  
        \includegraphics[height=6cm]{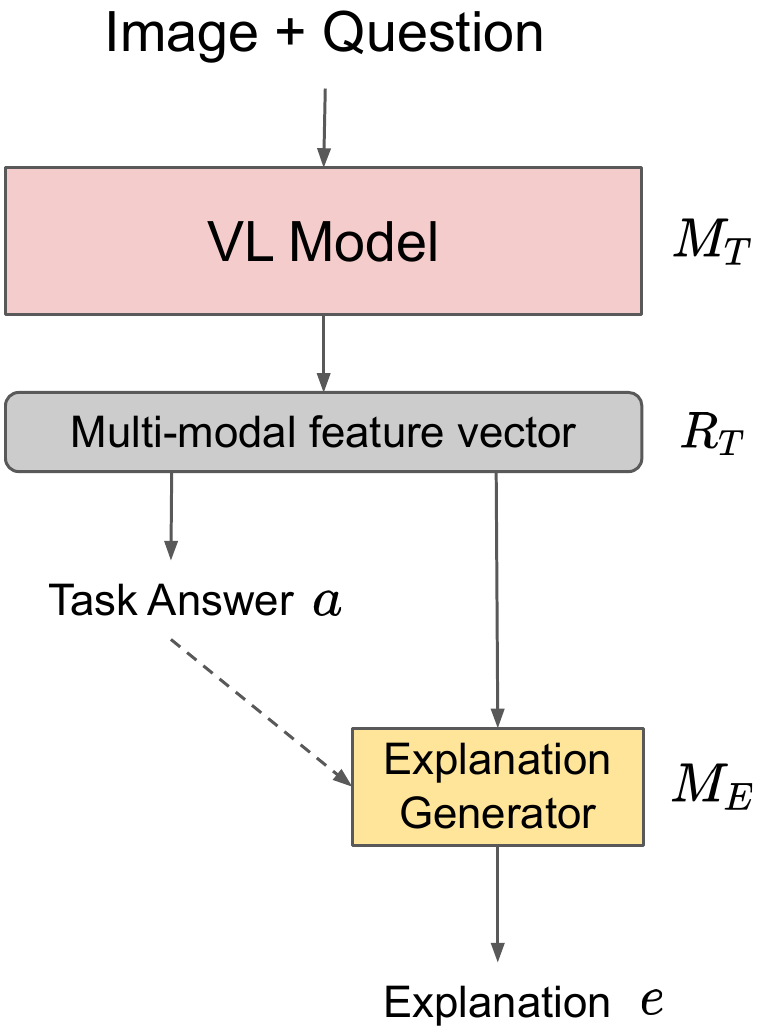}%
        \caption{High-level structure of VL models.}
    \end{subfigure}
\hfill
    \begin{subfigure}[h]{0.6\linewidth}
        \centering
        \includegraphics[height=6cm]{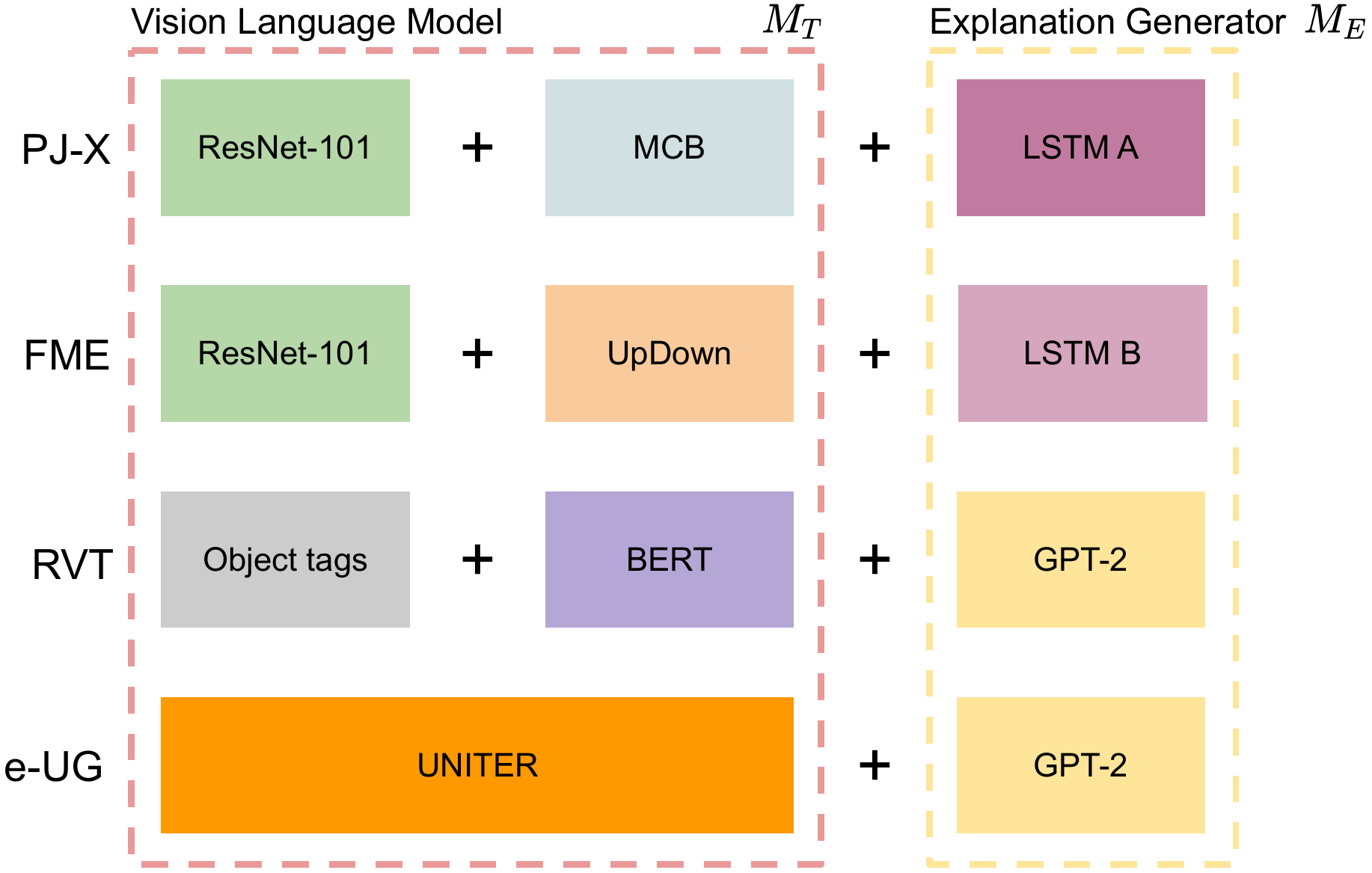}%
        \caption{The components of the models that we evaluate.}
    \end{subfigure}%
\caption{High-level architectures of the models that are included in our benchmark.}%
\label{fig:comArch}
\vspace{-2ex}
\end{figure*}

%% file: sections/experiments.tex
\subsection{Models}\label{models}

Existing VL-NLE models follow a common high-level structure (Figure \ref{fig:comArch}). First, a VL model learns a joint representation of the image and language inputs and predicts the answer. The models in this work then condition their explanation on different combinations of the question, image, their joint representation, and the answer. Details on PJ-X~\cite{park_multimodal_2018}, FME~\cite{wu_faithful_2019}, and RVT~\cite{marasovic_natural_2020} are given in Appendix~\ref{app:models}, as well as in their respective papers.

%The $M_T$ models in this work are MCB \cite{fukui2016multimodal}, UpDown \cite{anderson_bottom-up_2018}, BERT \cite{devlin2018bert} with object tags, and UNITER \cite{chen_uniter_2020}. The decoding of explanations is done via LSTM \cite{hochreiter1997long} architectures or the GPT-2 pre-trained language model \cite{radford2019language}.

%that we evaluate in this work are described below and are illustrated in Figure \ref{fig:comArch}. 

%The main differences between the models are outlined in Table \ref{tab:modDiff}.

\vspace{-2.5ex}
\paragraph{e-UG.}

\citet{marasovic_natural_2020} generate convincing explanations, but out of various $M_T$ modules tested, including complex visual reasoning models, it obtains the best explanation accuracy when using object labels as the sole image information. We address this limitation by proposing e-UG, a model that enables a stronger image conditioning by combining GPT-2 with UNITER~\cite{chen_uniter_2020}, a powerful transformer-based VL model.
%However, in many cases, object labels do not contain sufficient information to explain the answer to the question. This suggests that GPT-2 is able to learn explanations mostly from the questions and answers alone. To condition more strongly on the image, and at the same time use the general knowledge and fluency of GPT-2, we propose a novel model that combines GPT-2 with UNITER~\cite{chen_uniter_2020}, a powerful transformer based VL model. 
%Similar to BERT, UNITER leverages self-attention mechanisms to learn contextualized embeddings of image-text pairs. % \cite{chen_self-supervised_2019}. 
%It uses tasks such as Masked Language Modelling, Masked Region Modelling, Image-Text Matching, and Word-Region Alignment to pre-train on large image captioning datasets.
The outputs of UNITER are contextualized embeddings of the word tokens and image regions in the image-text pair. Words are embedded by tokenizing them into WordPieces and adding their position embedding. Images are embedded by extracting visual features of regions with Faster R-CNN \cite{ren2015faster} and encoding their location features. UNITER achieves SOTA on many downstream tasks when fine-tuned on them. For e-UG, we leverage these contextualized embeddings to condition GPT-2 on an efficient representation of the image and question. The embeddings of the image regions and question words are simply prepended to the textual question and predicted answer, and then fed to GPT-2. GPT-2 is a decoder-only architecture that is pre-trained on conventional language modeling and therefore well-suited for language generation~\cite{radford2019language}. We follow \citet{marasovic_natural_2020} and do greedy decoding during inference.

%\vspace{-2ex}
\subsection{Training}

\begin{table*}[htbp]
    \begin{center}
    \begin{tabulary}{\linewidth}{LCCCCCCCCCC}
    \toprule
        & Overall  & \multicolumn{3}{c}{VQA-X} & \multicolumn{3}{c}{e-SNLI-VE} & \multicolumn{3}{c}{VCR} \\
          \cmidrule(r){2-2} \cmidrule(r){3-5} \cmidrule(r){6-8} \cmidrule(r){9-11}
          & $S_E$         & $S_O$         & $S_T$         & $S_E$         & $S_O$         & $S_T$         & $S_E$         & $S_O$         & $S_T$ & $S_E$ \\
    \midrule
    PJ-X  & 59.2          & 49.9          & 76.4          & 65.4          & 41.2          & 69.2          & 59.6          & 20.6          & 39.0  & 52.7          \\
    FME & 60.1          & 47.7          & 75.5          & 63.2          & 43.1          & 73.7          & 58.5          & 28.6          & 48.9  & 58.5          \\
    RVT   & 62.8          & 46.0          & 68.6          & 67.1          & 42.8          & 72.0          & 59.4          & 36.4          & 59.0  & 61.8          \\
    e-UG  & \textbf{68.5} & \textbf{57.6} & \textbf{80.5} & \textbf{71.5} & \textbf{54.8} & \textbf{79.5} & \textbf{68.9} & \textbf{45.5} & \textbf{69.8}  & \textbf{65.1} \\
    GT & 79.3          & \multicolumn{1}{l}{--} & \multicolumn{1}{l}{--} & 84.5          &   --            &     --          & 76.2          &           --    &    --   & 77.3  \\
    \bottomrule
    \end{tabulary}\vspace{-1ex}
  \caption{e-ViL benchmark scores. $S_O$, $S_T$, and $S_E$ are defined in Section~\ref{sec:tf}. GT denotes the ground-truth explanations in each dataset. The best results are in bold.}%
  \label{tab:he_score}
  \end{center}
  \vspace{-4ex}
\end{table*}

All models are trained separately on each dataset. %This means we have a total of 12 model-dataset combinations. 
To ensure comparability, image features for PJ-X and FME are obtained from the same ResNet-101~\cite{he2016deep} pre-trained on ImageNet, which yields a 2048d feature representation for an image. % with 7x7 spatial dimension. 
To account for the small size of VQA-X, the VQA $M_T$ models were pre-trained on VQA v2 for VQA-X, and trained from scratch for the other two datasets. For UNITER, we follow the pre-training procedures used in the original paper~\cite{chen_uniter_2020}. The object tags in RVT are obtained from a Faster R-CNN that was trained on ImageNet and COCO. For GPT-2, we load the pre-trained weights of the original GPT-2 with 117M parameters~\cite{radford2019language}. For all models in this work, we experimented with training the $M_T$ and $M_E$ modules jointly and separately. More details are given in Appendix~\ref{app:jointSep}.

\vspace{-2.5ex}
\paragraph{Hyperparameters.}

Choosing hyperparameters via human evaluation is prohibitively expensive. Instead, we defined a set of automatic NLG metrics that we used to approximate the selection of the best hyperparameters. We define the score of an explanation as the harmonic mean of the BERTScore F1~\cite{zhang2019bertscore} and $\mathrm{NGRAMScore}$, where we set $\mathrm{NGRAMScore}$ as the harmonic mean of the $n$-gram NLG metrics ROUGE-L~\cite{lin2004automatic}, SPICE~\cite{anderson2016spice}, CIDEr~\cite{vedantam2015cider}, and METEOR~\cite{banerjee2005meteor}. We pick the harmonic mean, as it puts more emphasis on the weaker scores. Further details on the hyperparameters are given in Appendix \ref{app:hyp}.

% FOR PJX/FM on VCR and esnlive:  We pick as as final candidate the model minimising val_accracy * $m_e$.
% For PJX/FM we trim our dictionary ro only include words occuring 3 or more times across any the dataset resultsing in Dict sizes for VCR: 20782 words and esnlive: 19456 words

\subsection{Results}

In this section, we highlight the human evaluation results, their correlation with automatic NLG metrics, and the effect that training with explanations has on the performance on task $T$.
%provide the results obtained by the different models for human evaluation. We also provide a study on the correlation between different automatic NLG metrics and human evaluation scores. Lastly, we look at the effect that training with explanations has on the performance on task $T$. 
Model performance for automatic NLG metrics, detailed results on e-SNLI-VE, alternative computations of the human evaluation score, and a statistical analysis of the results is provided in Appendix \ref{app:results}.

\vspace{-2ex}
\subsubsection{Human Evaluation}
\begin{figure}
\begin{center}
\includegraphics[width=1\linewidth]{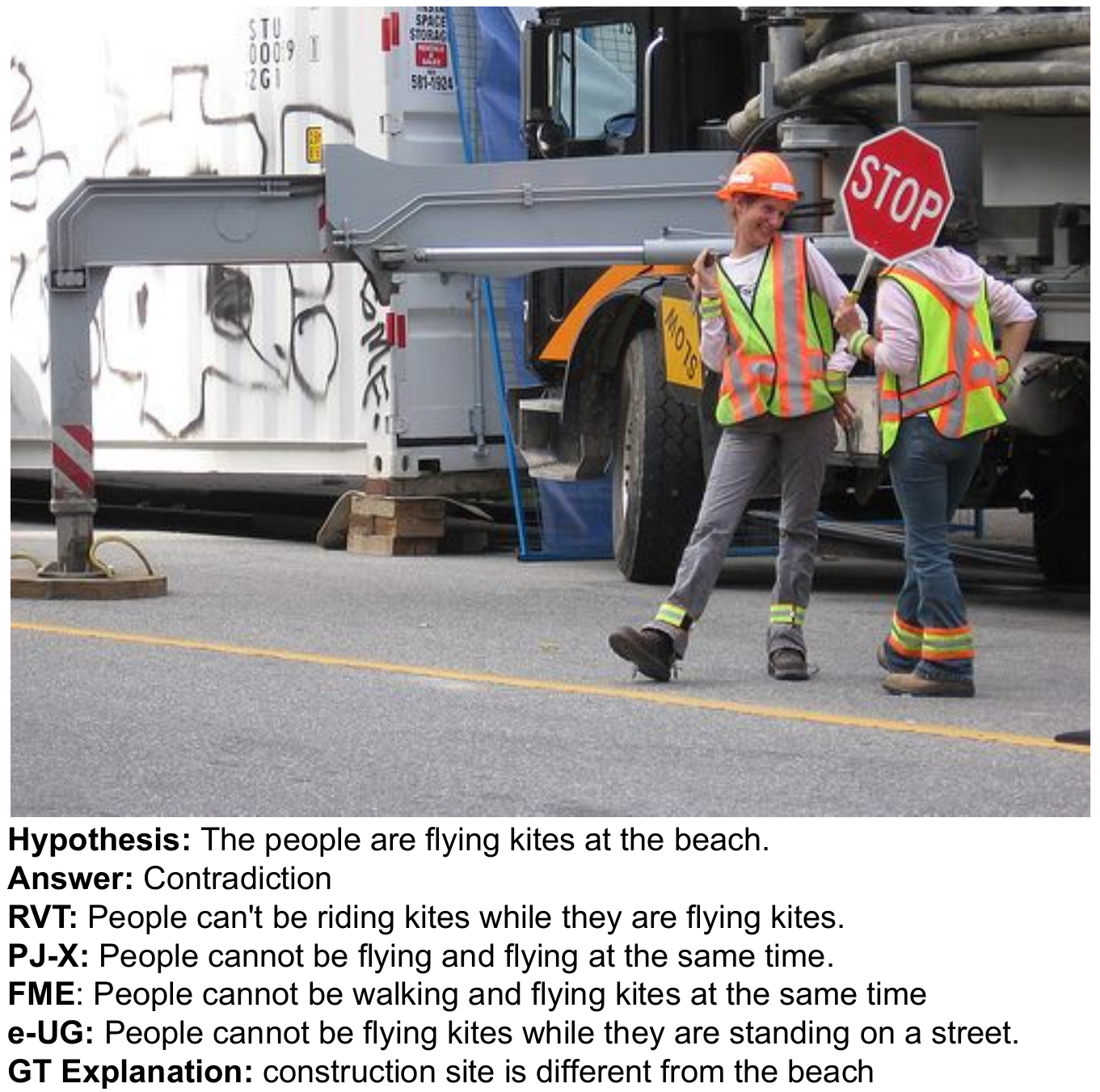}
\end{center}\vspace{-1.5ex}
    \caption{Generated explanations for each model on an image-hypothesis pair in e-SNLI-VE.}%
    \label{fig:test_ex}
\vspace{-2ex}
\end{figure}

The explanation scores $S_E$ obtained from the e-ViL human evaluation framework are displayed in Table~\ref{tab:he_score}. Our model e-UG outperforms existing methods on all datasets, with an average $S_E$ score 5.7 points higher than the second-best model, RVT. Despite leveraging little image information, RVT achieves higher scores than  PJ-X and FME on average, reflecting the ability of GPT-2 to learn to generate convincing explanations, without much anchoring on the image. There is still a significant gap between $S_E$ scores of generated explanations and ground-truth (GT) explanations. For VQA-X, $S_E$ scores are higher overall, indicating that the dataset is easier. In terms of the overall score $S_O$, the gap between e-UG and the rest increases further, as UNITER achieves a higher performance on VL tasks than the $M_T$ modules of the other models. In Figure \ref{fig:test_ex}, we show an example with the explanations generated by each model. In this example, e-UG is the only model that accurately describes the image and justifies the answer. Additional examples are given in Figure \ref{fig:gen_examples} in the appendix. 

As a second question, we ask the annotators to select the shortcomings (if any) for every explanation. Results for this are given in Table~\ref{tab:shortcoming}. The most frequent shortcoming is an insufficient justification of the answer. Least frequent, with around 10\% prevalence, explanations can be nonsensical (e.g., ``a woman is a woman"). All models struggle similarly much with producing explanations that sufficiently justify the answer. e-UG and PJ-X are better at producing coherent sentences. e-UG is significantly superior in terms of the explanations accurately describing the image content. This empirically confirms the effectiveness of our enhanced conditioning on the image. On a dataset level, we see that it is easiest for all models to provide explanations that make grammatical sense and justify the answer on VQA-X, reinforcing the notion that the explanations of VQA-X are easier and less elaborate.

A statistical analysis of our findings are given in Appendix~\ref{app:results}.

\begin{table*}[h!]
    \begin{center}
    \begin{tabulary}{\linewidth}{LLCCCCCC}
    \toprule
                 &  & \multicolumn{2}{c}{VQA-X}      & \multicolumn{2}{c}{SNLI-VE}           & \multicolumn{2}{c}{VCR}               \\
    \cmidrule(r){3-4} \cmidrule(r){5-6} \cmidrule(r){7-8} 
    Model       & $M_T$ model   & $M_T$ only & Joint  & $M_T$ only        & Joint & $M_T$ only        & Joint \\
    \mbox{PJ-X} & MCB~\cite{fukui2016multimodal}     & N.A.          & N.A.           & \underline{69.7} & 69.2           & 38.5 & \underline{39.0}           \\
    \mbox{FME}& UpDown~\cite{anderson2018bottom} & N.A.          & N.A.           & 71.4 & \underline{73.7}           & 35.7 & \underline{48.9}           \\
    e-UG & UNITER~\cite{chen_uniter_2020}  & 80.0          & \underline{80.5}           & 79.4                 & 79.5           & 69.3                 & \underline{69.8}           \\
    \bottomrule
    \end{tabulary}\vspace{-1ex}
  \caption{Comparison of task scores $S_T$ (e.g., accuracies) when the models are trained only on task $T$ vs.\ when trained jointly on tasks $T$ and $E$. Scores are underlined if their difference is greater than 0.5.}%
  \label{tab:bb}
  \end{center}\vspace{-4ex}
\end{table*}

\begin{table}[htbp!]
    \begin{center}
    \begin{tabulary}{\linewidth}{LCCCC}
    \toprule
    Model   & Untrue to Image & Lack of Justification & Non-sensical Sentence \\
    \midrule
    \mbox{PJ-X}  & 25.0\%          & 26.4\%          & 8.9\%          \\
    RVT   & 20.4\%          & 24.2\%          & 12.0\%         \\
    \mbox{FME} & 21.8\%          & \textbf{23.1\%} & 13.7\%         \\
    \mbox{e-UG}  & \textbf{15.9\%} & 25.0\%          & \textbf{7.4\%} \\
    \midrule
    Dataset   &  &  &  \\
    \midrule
    \mbox{e-SNLI-VE} & 21.3\%          & 28.7\%                & 12.8\%                \\
    VCR       & 21.0\%          & 31.2\%                & 11.7\%                \\
    \mbox{VQA-X}     & 20.0\%          & 15.4\%                & 7.4\%                 \\
    \bottomrule
    \end{tabulary}\vspace{-1ex}
  \caption{Main shortcomings of the generated explanations, by models and by datasets. Human judges could choose multiple shortcomings per explanation. The best model results are in bold.}%
  \label{tab:shortcoming}
  \end{center}
  \vspace{-4ex}
\end{table}

\vspace{-2ex}
\subsubsection{Correlation of NLG Metrics with Human Evaluation}

To better understand to what extent automatic NLG metrics are able to mirror human judgment of explanations, we compute the Spearman correlation of different NLG metrics with the human evaluation scores. The NLG metrics for the different models are given in Appendix~\ref{app:sec:autoNLG}.
The human evaluation score is averaged and normalised (across all annotators) for each explanation. We have human evaluation scores for a total of  3,566\footnote{We have 4 models, 3 datasets of 300 examples, therefore 3,600 explanations. However, for 34 of them, all the three annotators answered the question incorrectly.} generated explanations, which makes it the currently largest study on the correlation of NLG metrics with human evaluation in NLEs. 

The results in Table~\ref{tab:corr} show that BERTScore and METEOR exhibit significantly higher correlation with human annotators across all datasets, reaching a maximal value of 0.293, which is a relatively low correlation. The reliability of automatic metrics also differs by dataset. They are highest on VQA-X and lowest on VCR. This could be explained by the fact that explanations in VCR are generally semantically more complex or more speculative (and, therefore, there are more different ways to explain the same thing) than those in VQA-X. It is noteworthy that some $n$-gram metrics, such as BLEU, ROUGE, or CIDEr, have no statistically significant correlation with human judgment on VCR. 

\begin{table}[htbp]
    \begin{center}
    \begin{tabulary}{\linewidth}{LCCCC}
    \toprule
    Metric    & All datasets & VQA-X & e-SNLI-VE & VCR            \\
    \midrule
    BLEU-1    & 0.222        & 0.396 & 0.123     & \textit{0.032} \\
    BLEU-2    & 0.236        & 0.412 & 0.142     & \textit{0.034} \\
    BLEU-3    & 0.224        & 0.383 & 0.139     & \textit{0.039} \\
    BLEU-4    & 0.216        & 0.373 & 0.139     & \textit{0.038} \\
    METEOR    & 0.288        & \textbf{0.438} & 0.186     & 0.113          \\
    \mbox{ROUGE-L}   & 0.238        & 0.399 & 0.131     & \textit{0.050} \\
    CIDEr     & 0.245        & 0.404 & 0.133     & \textit{0.093} \\
    SPICE     & 0.235        & 0.407 & 0.162     & 0.116          \\
    BERTScore & \textbf{0.293}        & 0.431 & 0.189     & \textbf{0.138}          \\
    BLEURT~\cite{sellam2020bleurt}    & 0.248        & 0.338 & \textbf{0.208}     & 0.128        \\
    \bottomrule
    \end{tabulary}\vspace{-1ex}
  \caption{Correlation between human evaluation and automatic NLG metrics on NLEs. All values, except those in \textit{italic}, have p-values $< 0.001$.}%
  \label{tab:corr}
  \end{center}
  \vspace{-4ex}
\end{table}

\vspace{-1.5ex}
\subsubsection{Explanations as Learning Instructions} \label{sec:bb}
\vspace{-0.5ex}
Training a model jointly on the tasks $T$ and $E$ can be viewed as a form of multi-task learning~\cite{caruana1997multitask}. The explanations $e$ augment the datapoints of task $T$ by explaining why an answer $a$ was given. The module $M_T$ (which solves task $T$) may benefit from this additional signal. Indeed, the model is forced to learn a representation of the image and question from which both the answer and explanation can be extracted, which could improve the model's representation capabilities. To verify this hypothesis, we compare the task scores of modules $M_T$ that trained only on task $T$ and those that, together with $M_E$, were jointly trained on tasks $T$ and $E$. We do this for e-UG on all three datasets, and for FME and PJ-X on VCR and e-SNLI-VE (because a larger pre-training dataset exists for VQA-X). The results in Table~\ref{tab:bb} show that, without any adaptations, the task performance for joint training is equal or better in all but one model-dataset combination. These results suggests that explanations may have the potential to act as ``learning instructions" and thereby improve the classification capabilities of a model. Additional experiments are required to further verify this and to develop approaches that more efficiently leverage the explanations.

\vspace{-0.5ex}

%% file: sections/conclusion.tex
\vspace{-0.5ex}
We addressed the lack of comparison between existing VL-NLE methods by introducing e-ViL, a unified and re-usable benchmark on which we evaluated four different architectures using human judges. We also introduced e-SNLI-VE, the largest existing VL dataset with human-written explanations. The e-ViL benchmark can be used by future works to compare their VL-NLE models against existing ones. Furthermore, our correlation study has shown that automatic NLG metrics have a weak correlation with human judgment. In this work, we also propose a new model, e-UG, which leverages contextualized embeddings of the image-question pairs and achieves a state-of-the-art performance by a large margin on all datasets. 

Important questions that need to be addressed in future work are the faithfulness of the explanations (i.e., that they faithfully reflect the model reasoning) and the development of automatic NLG metrics that have a stronger correlation with human judgment.

\vspace{-0.5ex}

%% file: sections/acknowledge.tex
\vspace{-0.5ex}
Maxime Kayser, Leonard Salewski, and Cornelius Emde are supported by Elsevier BV, the International Max Planck Research School for Intelligent Systems, and by Cancer Research UK (grant number C2195/A25014), respectively. This work has been partially funded by the ERC (853489---DEXIM) and by the DFG (2064/1---Project number 390727645). This work has also been supported by the Alan Turing Institute under the EPSRC
grant EP/N510129/1,  by the AXA Research Fund,  the ESRC
grant “Unlocking the Potential of AI for English Law”,  the EPSRC grant EP/R013667/1, 
and by the EU TAILOR grant. We also acknowledge the use of Oxford’s Advanced Research Computing (ARC) facility, of the EPSRC-funded Tier 2 facility
JADE (EP/P020275/1), and of GPU computing support by
Scan Computers International Ltd.

%Leonard Salewski is supported by the International Max Planck Research School for Intelligent Systems. Cornelius Embde. The authors thank the International Max Planck Research School for Intelligent Systems (IMPRS-IS) for supporting Leonard Salewski. This work was supported by Cancer Research UK (CRUK), through a CRUK Oxford Centre Prize DPhil Studentship .
\cleardoublepage

%% file: appendix/examples.tex
\begin{figure*}[ht!] 
    \begin{subfigure}[h]{0.5\linewidth}
        \centering  
        \includegraphics[height=0.3\textheight]{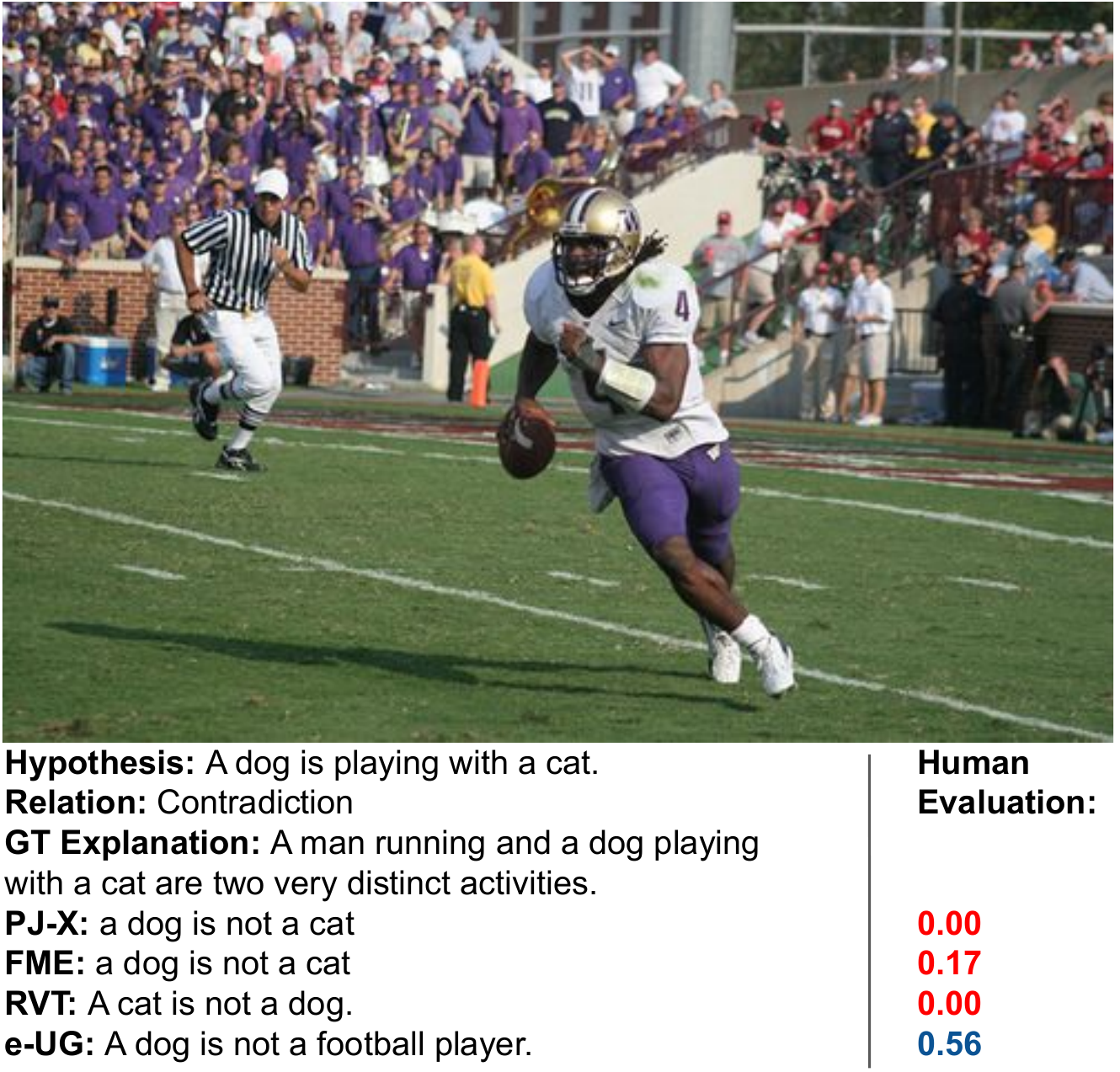}%
        \caption{e-SNLI-VE.} \label{fig:1a}
    \end{subfigure}
\hfill
    \begin{subfigure}[h]{0.5\linewidth}
        \centering
        \includegraphics[height=0.3\textheight]{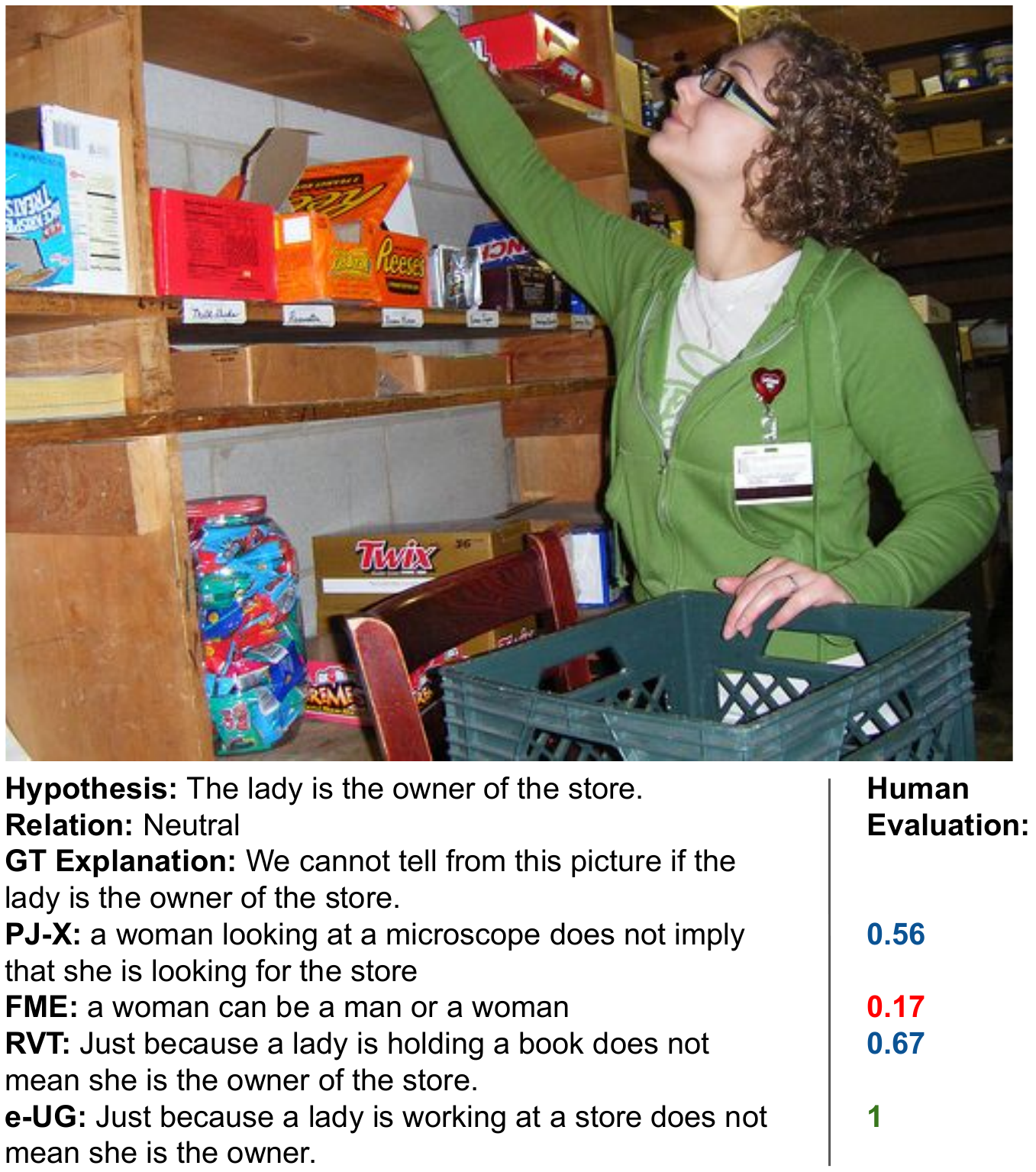}%
        \caption{e-SNLI-VE.} \label{fig:1b}
    \end{subfigure}
\hfill
    \begin{subfigure}[h]{0.5\linewidth}
        \centering
        \includegraphics[height=0.3\textheight]{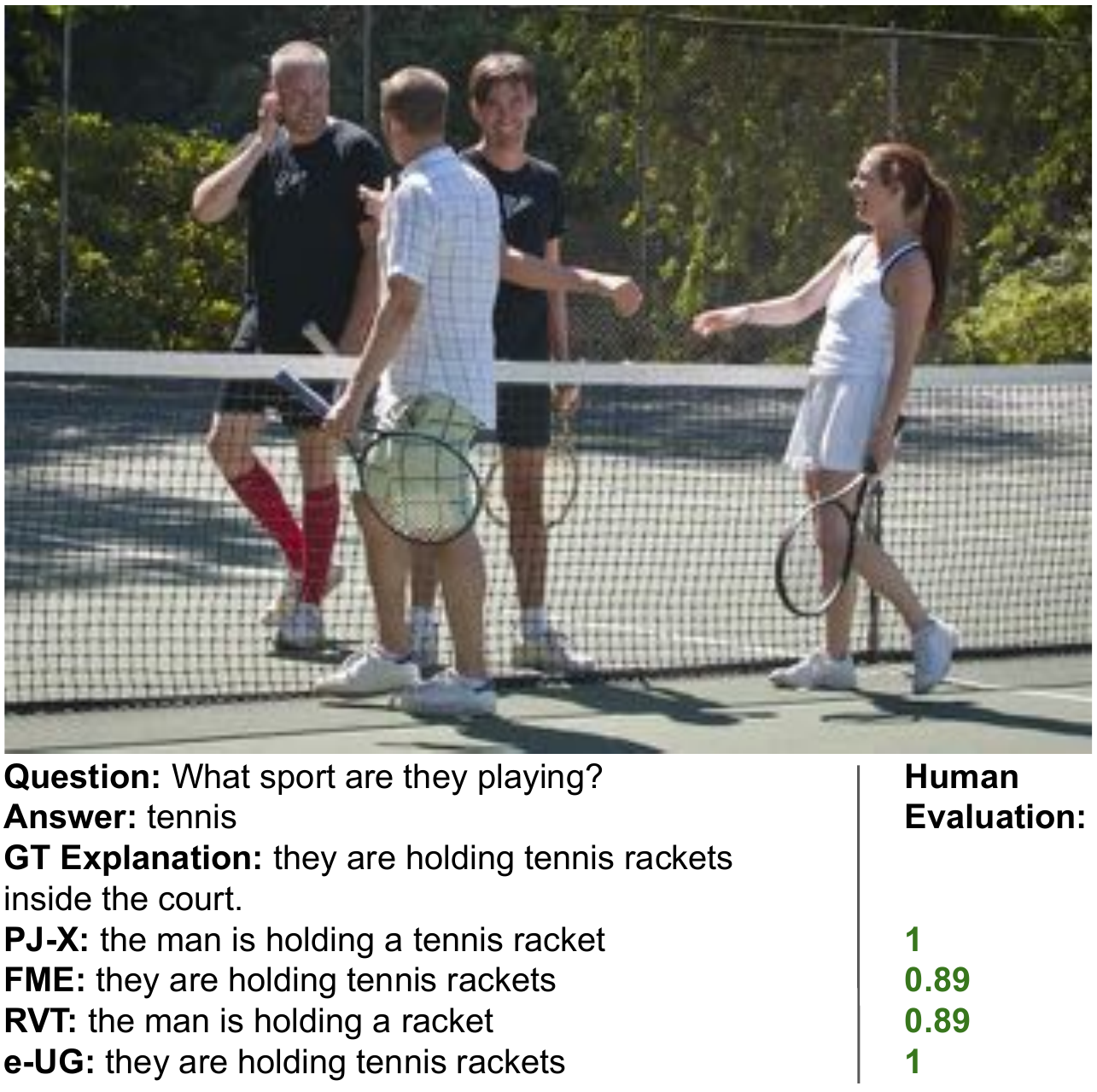}%
        \caption{VQA-X.} \label{fig:1c}
    \end{subfigure}%
\hfill
    \begin{subfigure}[h]{0.5\linewidth}
        \centering
        \includegraphics[height=0.3\textheight]{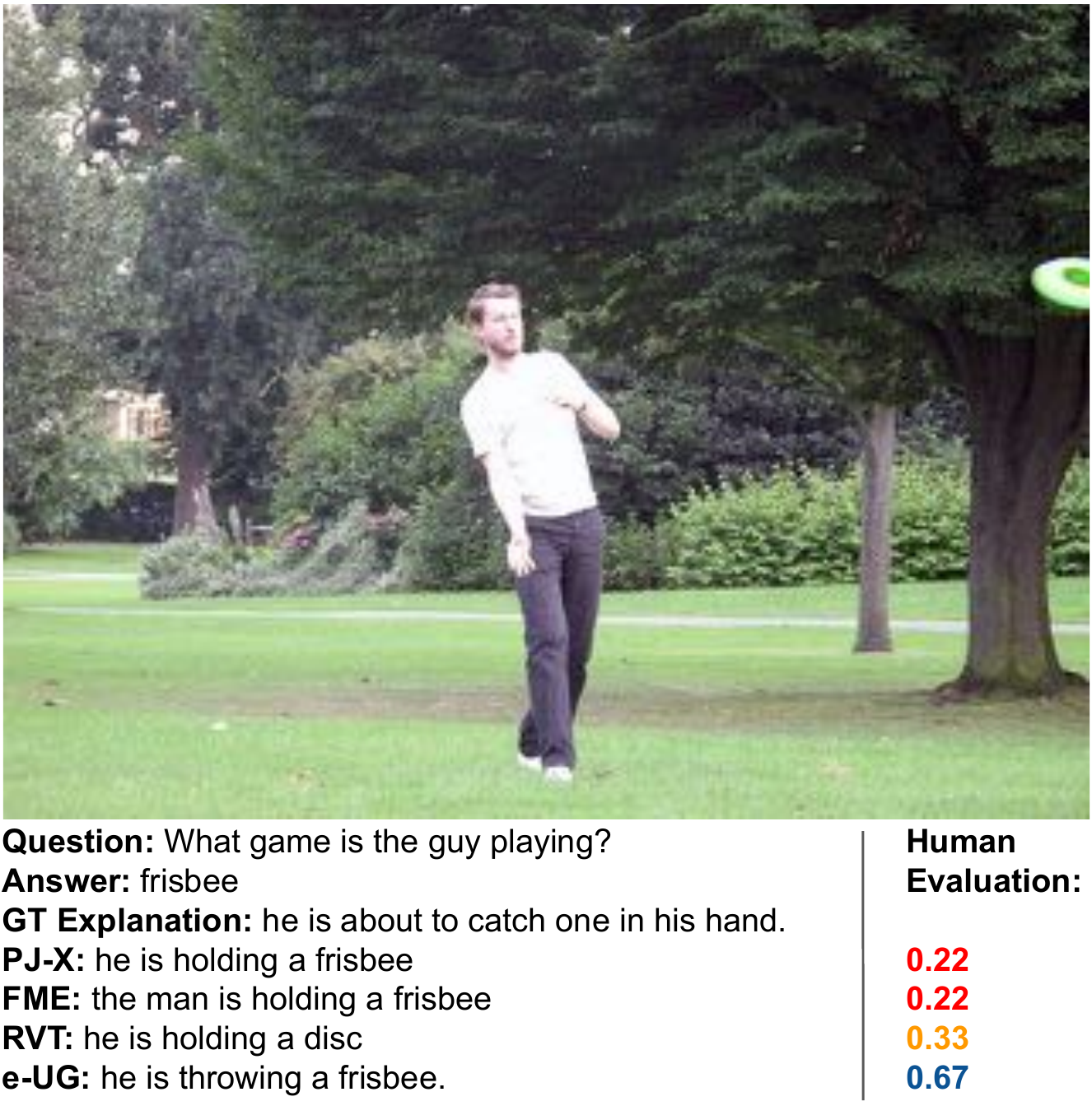}%
        \caption{VQA-X.} \label{fig:1d}
    \end{subfigure}%
\hfill
    \begin{subfigure}[h]{0.5\linewidth}
        \centering
        \includegraphics[height=0.3\textheight]{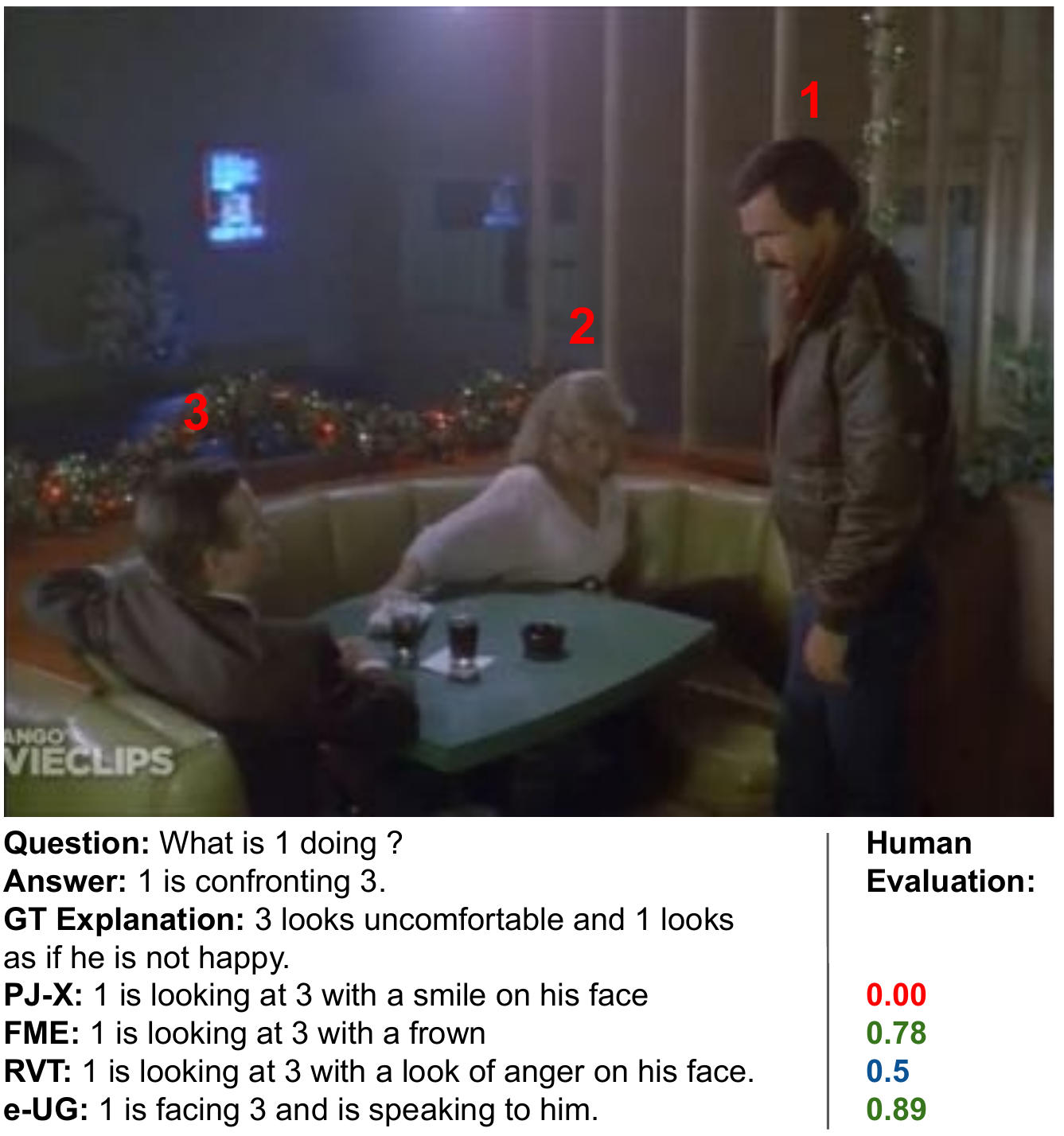}%
        \caption{VCR.} \label{fig:1e}
    \end{subfigure}%
\hfill
    \begin{subfigure}[h]{0.5\linewidth}
        \centering
        \includegraphics[height=0.3\textheight]{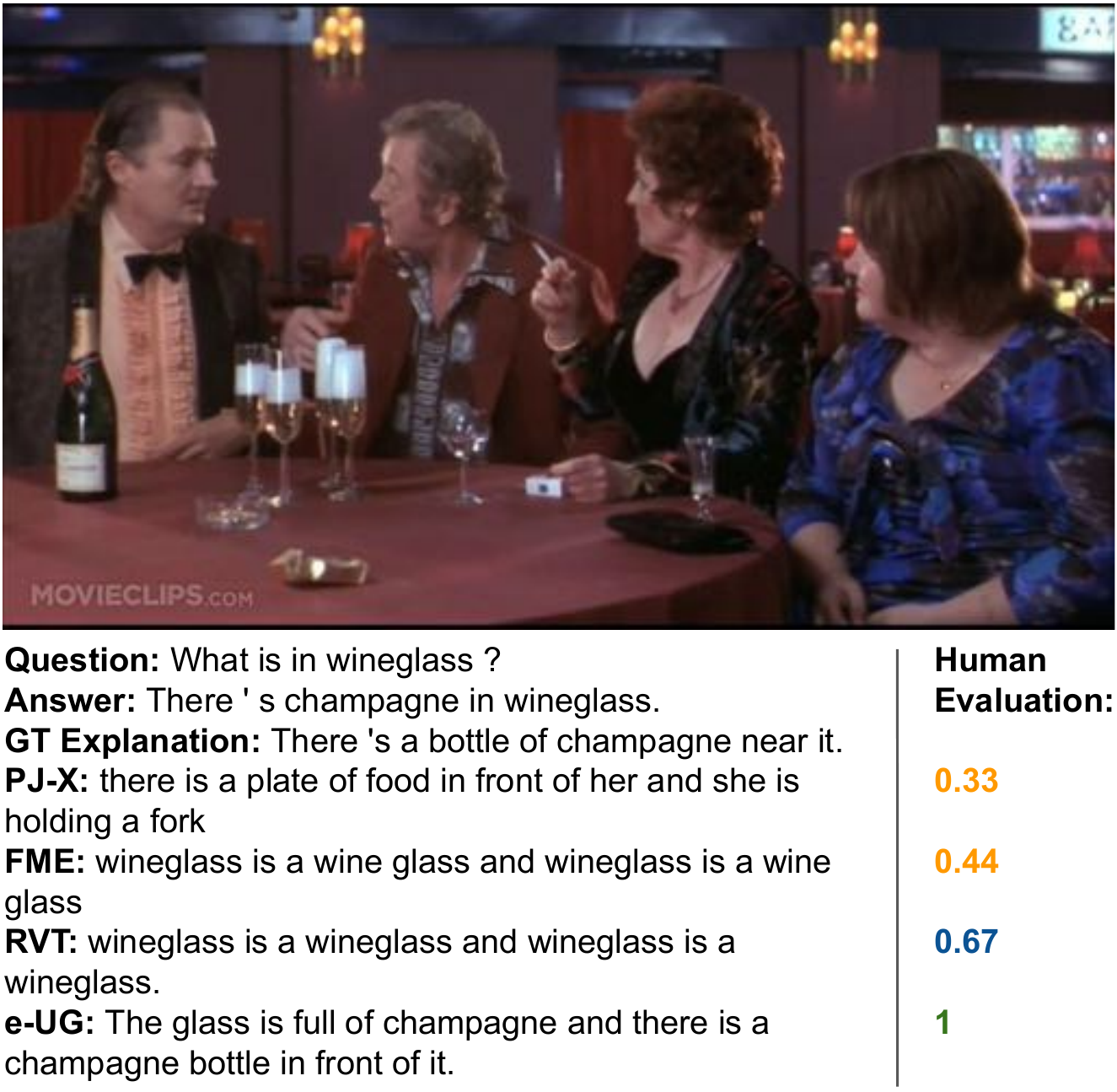}%
        \caption{VCR.} \label{fig:1f}
    \end{subfigure}%
\caption{Pair of examples from the test set of each dataset. We display the ground-truth (GT) explanation, as well as the generated explanations of each model and their predicted human evaluation score $S_E$.}%
\label{fig:gen_examples}
\vspace{-2ex}
\end{figure*}

Two sets of figures are given to show qualitative examples of our datasets and models. Figure~\ref{fig:gen_examples} shows the explanations generated by the models (and the ground-truth) for two images of each dataset. We  also display the e-ViL $S_E$ score of each generated explanation, which was obtained through our human evaluation framework. In some of the images, such as in Figures \ref{fig:1b} and \ref{fig:1f}, we can see that e-UG provides better, more image-grounded explanations. 

\begin{figure*}[ht!] 
    \begin{subfigure}[h]{0.5\linewidth}
        \centering  
        \includegraphics[height=6cm]{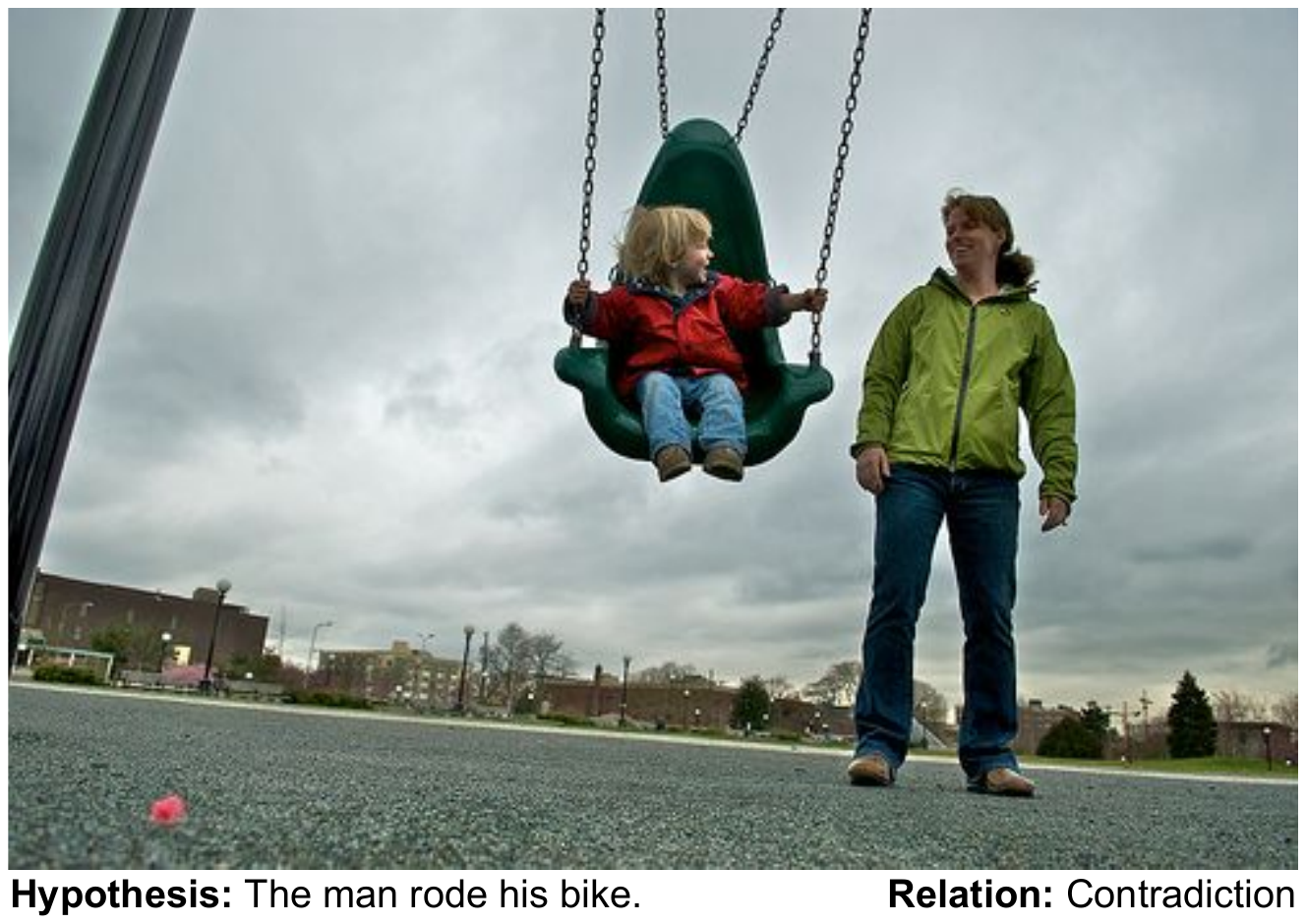}%
        \caption{e-SNLI-VE.} \label{fig:2a}
    \end{subfigure}
\hfill
    \begin{subfigure}[h]{0.5\linewidth}
        \centering
        \includegraphics[height=6cm]{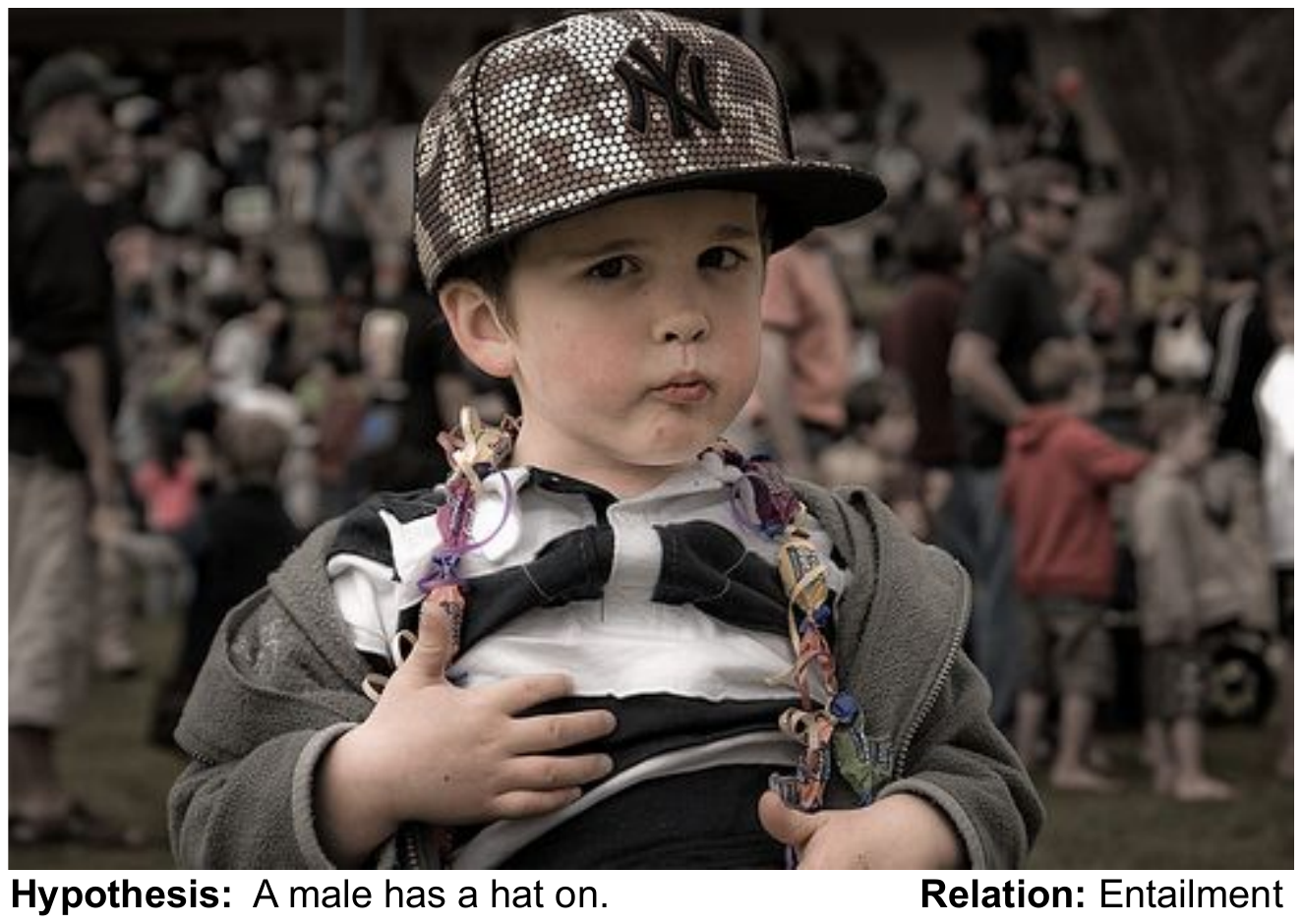}%
        \caption{e-SNLI-VE.} \label{fig:2b}
    \end{subfigure}%
\hfill
    \begin{subfigure}[h]{0.5\linewidth}
        \centering
        \includegraphics[height=6cm]{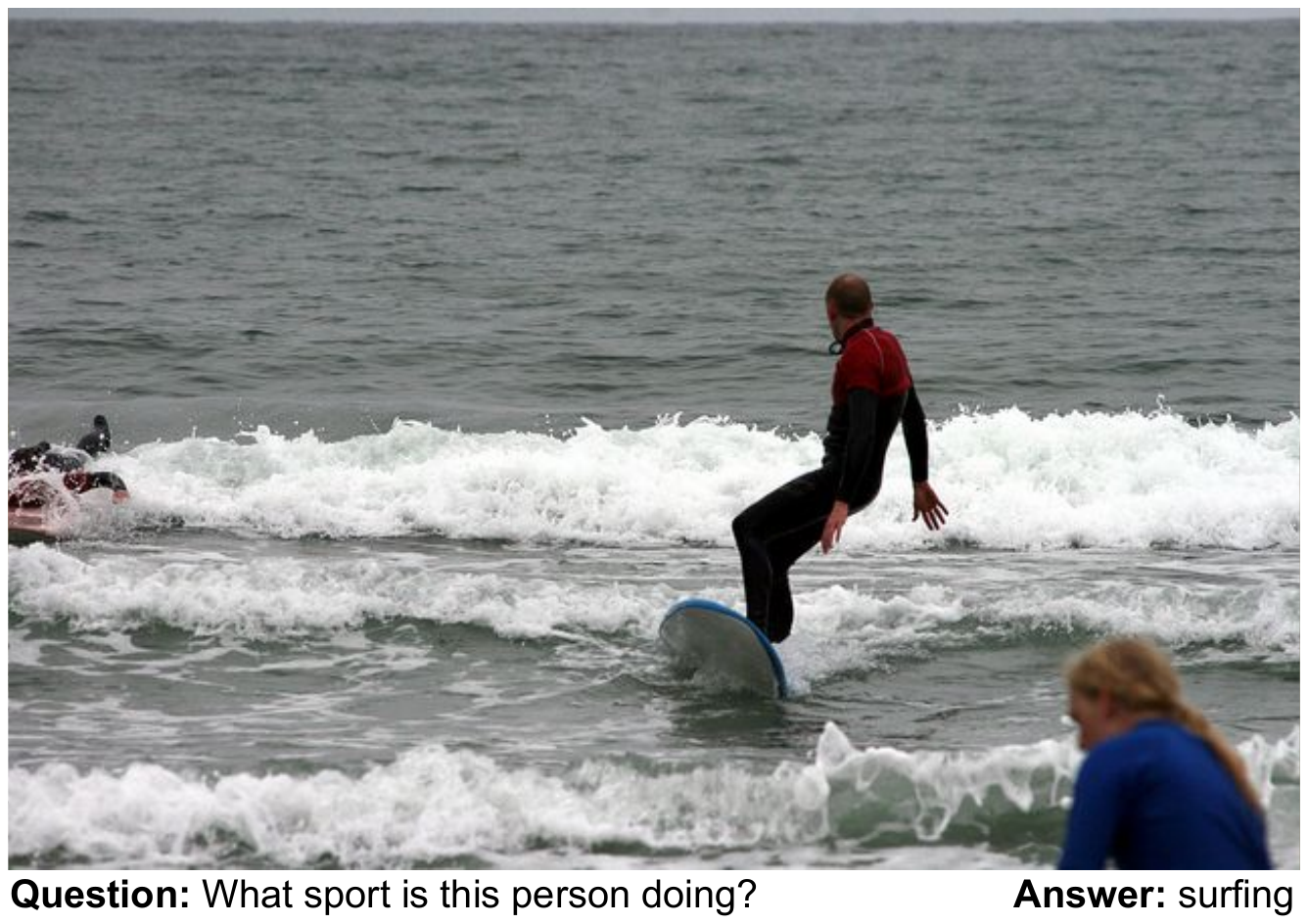}%
        \caption{VQA-X.} \label{fig:2c}
    \end{subfigure}%
\hfill
    \begin{subfigure}[h]{0.5\linewidth}
        \centering
        \includegraphics[height=6cm]{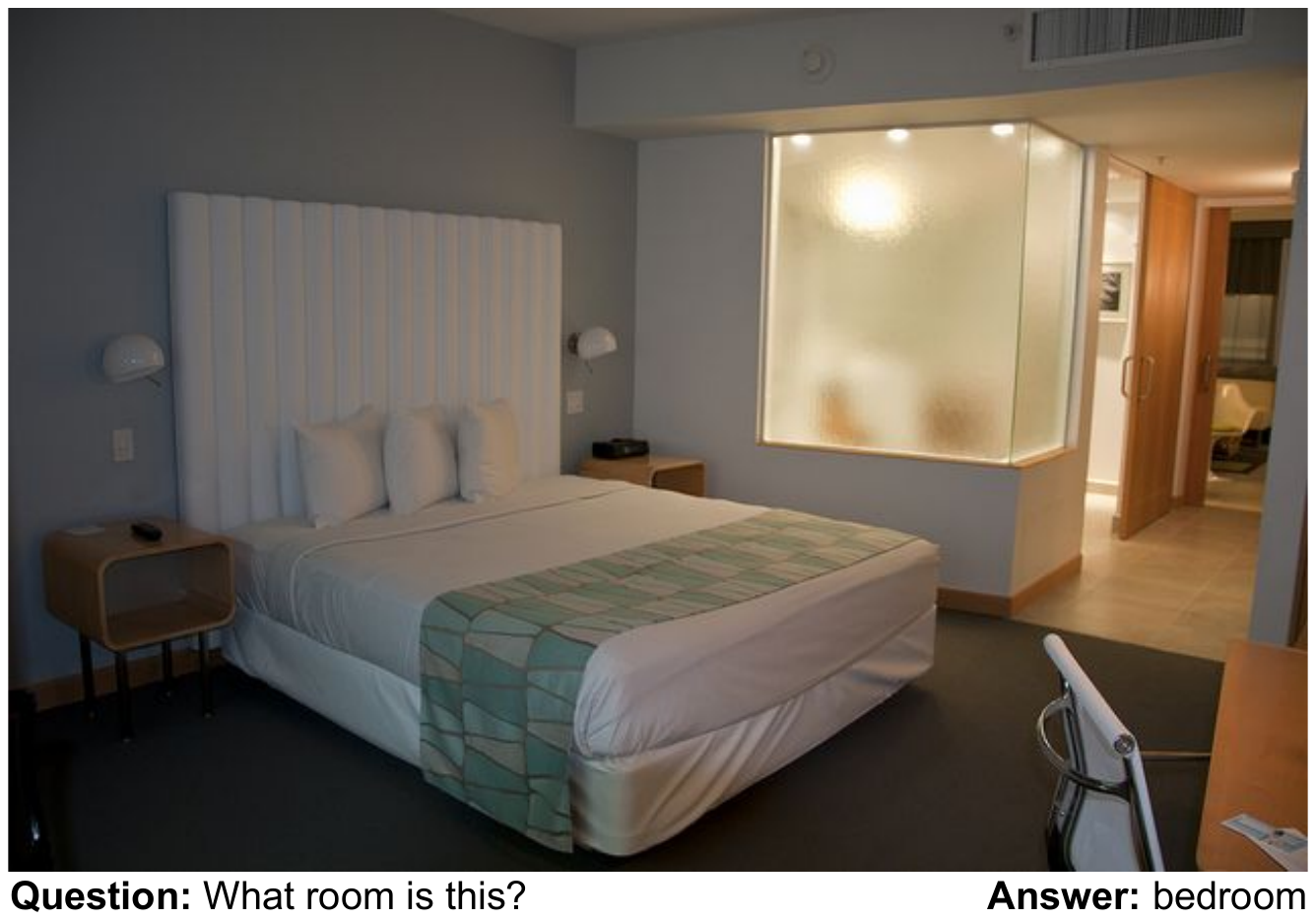}%
        \caption{VQA-X.} \label{fig:2d}
    \end{subfigure}%
\hfill
    \begin{subfigure}[h]{0.5\linewidth}
        \centering
        \includegraphics[height=6cm]{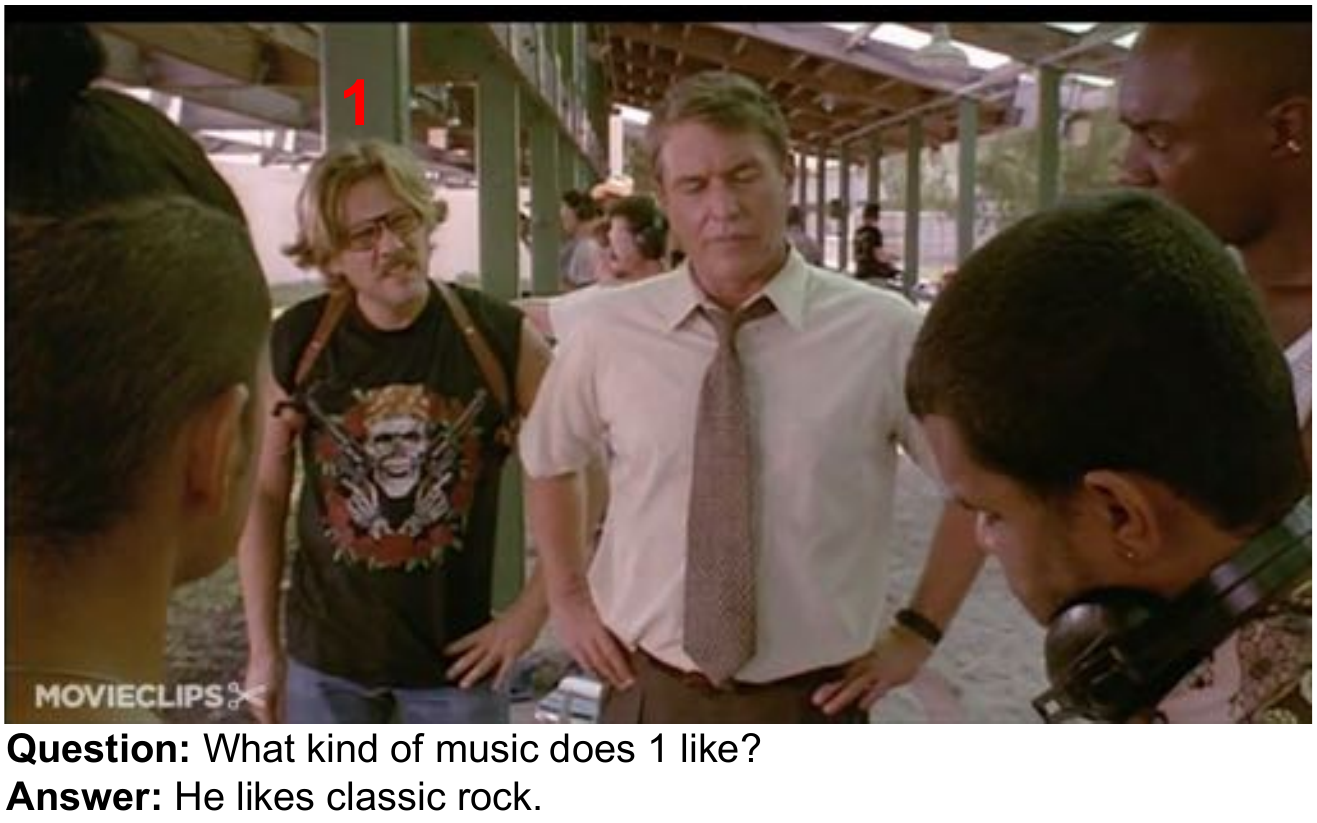}%
        \caption{VCR.} \label{fig:2e}
    \end{subfigure}%
\hfill
    \begin{subfigure}[h]{0.5\linewidth}
        \centering
        \includegraphics[height=6cm]{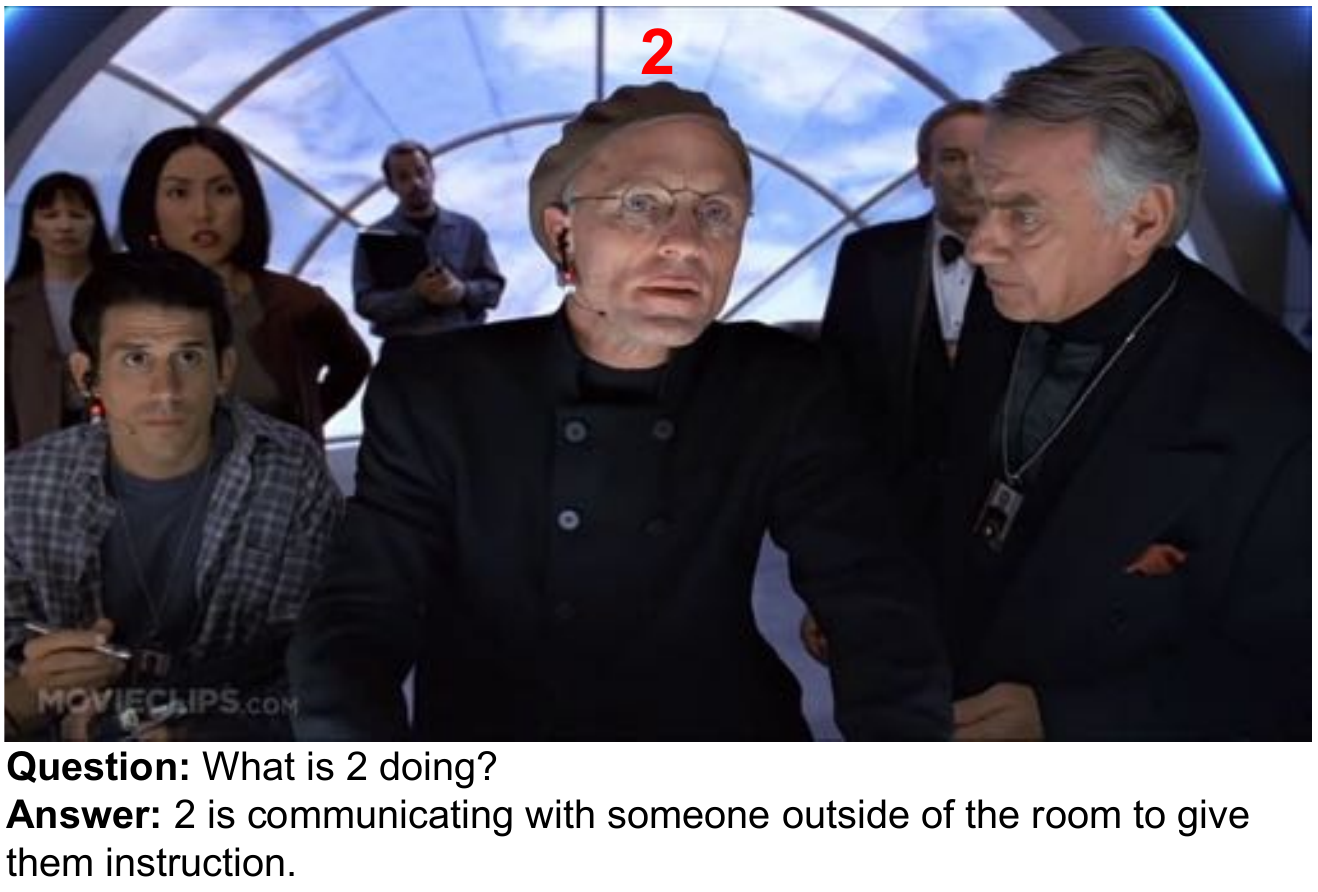}%
        \caption{VCR.} \label{fig:2f}
    \end{subfigure}%
\caption{Representative examples from each dataset.}%
\label{fig:dset_examples}
\vspace{-2ex}
\end{figure*}

In Figure~\ref{fig:dset_examples} we again show two images per dataset. These examples illustrate the key differences between the different datasets. VCR has many questions that require substantial commonsense reasoning and general knowledge. For example, to explain the answer to the question in Figure \ref{fig:2e}, one needs to know that person 1 is wearing a T-shirt of the classic rock band Guns N' Roses. For VQA-X (Figures \ref{fig:2c} and \ref{fig:2d}), we show two examples where a generic explanation, that is not necessarily grounded in the image, will often suffice (this is a general limitation of this dataset). The explanation ``Because there is a person on a surfboard" and ``Because there is a bed in the room" will, in most cases, be correct with respect to the question and answer, regardless of the image. The examples for e-SNLI-VE in Figures \ref{fig:2a} and \ref{fig:2b} both require the explanations to describe image-specific characteristics in order to be meaningful. In Figure \ref{fig:2a}, a valid explanation would have to pick out a concrete element from the image to explain why it is a contradiction.

%% file: appendix/eSNLIVE.tex
This section contains a datasheet on e-SNLI-VE, as well as further information on its pre-processing. Details on the filters are given in Section~\ref{sec:app_filt}. Details on the MTurk evaluation can be found in Appendix~\ref{app:mturk}. %This was wrongly referenced in the main paper. 

\subsection{e-SNLI-VE Datasheet}
\input{appendix/datasheet}

\subsection{Relabeling e-SNLI-VE via MTurk} \label{app:mturk_relab}

\begin{figure}[H]
\begin{center}
\includegraphics[width=1\linewidth]{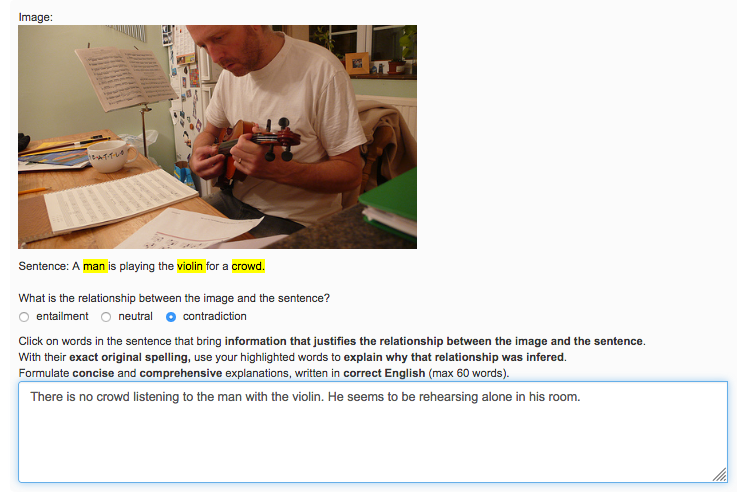}
\end{center}
   \caption{A snapshot of the annotation interface that was used to manually reannotate the neutral labels in the validation and test sets of e-SNLI-VE.}%
\label{fig:amt_esnlive_setup}
\vspace{-2ex}
\end{figure}

In this work, we collect new labels and explanations for the neutral pairs of the validation and test sets of e-SNLI-VE. We provide workers with the definitions of entailment, neutral, and contradiction for image-sentence pairs and one example for each label. As shown in Figure~\ref{fig:amt_esnlive_setup}, for each image-sentence pair, workers are required to (a) choose a label, (b) highlight words in the sentence that led to their label decision, and (c) explain their decision in a comprehensive and concise manner, using at least half of the words that they highlighted. We point out that it is likely that requiring an explanation at the same time as requiring a label has a positive effect on the correctness of the label, since having to justify in writing the picked label may make annotators pay an increased attention. Moreover, we implemented additional quality control measures for crowdsourced annotations, such as (a) collecting three annotations for every input, (b) injecting trusted annotations, and (c) restricting to annotators with at least 90\% previous approval rate. 

\begin{figure}[!h]
\begin{center}
\includegraphics[width=1\linewidth]{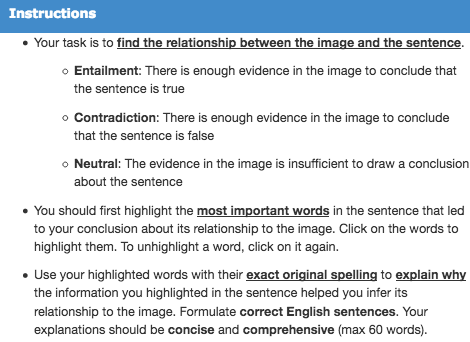}
\end{center}
   \caption{A snapshot of the instructions that were provided to the workers that reannotated the neutral labels in the validation and test sets of e-SNLI-VE.}%
\label{fig:amt_esnlive_inst}
\vspace{-2ex}
\end{figure}

%We used Amazon Mechanical Turk (MTurk) to collect new labels and explanations for SNLI-VE. 
There were 2,060 workers in the annotation effort, with an average of 1.98 assignments per worker and a standard deviation of 5.54. No restriction was put on the workers’ location. Each assignment consisted of a set of 10 image-sentence pairs. The instructions are shown in Figure~\ref{fig:amt_esnlive_inst}. The annotators were also guided by three examples, one for each label. For each assignment of 10 questions, one trusted annotation with known label was inserted at a random position, as a measure to control the quality of label annotation. Each assignment was completed by three different workers.

To check the success of our crowdsourcing, we manually assessed the relevance of explanations among a random subset of 100 examples. A marking scale between 0 and 1 was used, assigning a score of k/n when k required attributes were given in an explanation out of n. We report an 83.5\% relevance of explanations from workers.

\subsection{Ambiguity in e-SNLI-VE}

\begin{figure}[!h]
\centering
\includegraphics[scale=0.3]{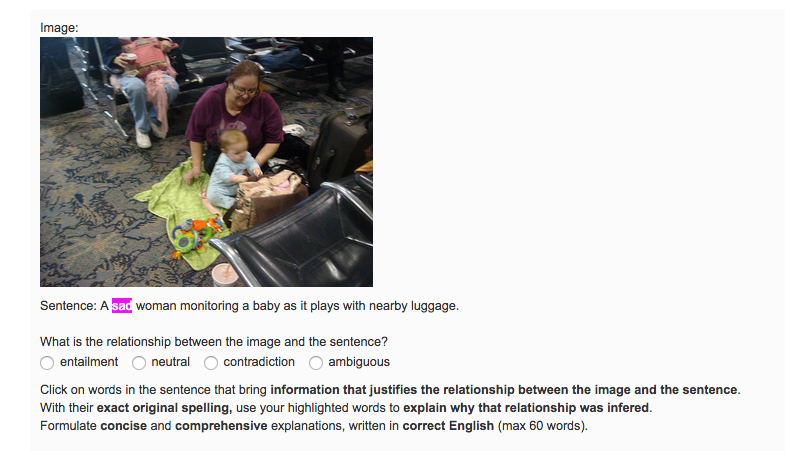}
\caption{\label{fig:ambiguous2}Ambiguous SNLI-VE instance. Some may argue that the woman's face betrays sadness, but the image is not quite clear. Secondly, even with better resolution, facial expression may not be a strong enough evidence to support the hypothesis about the woman's emotional state.}
\end{figure}

\begin{figure}[!h]
\centering
\includegraphics[scale=0.3]{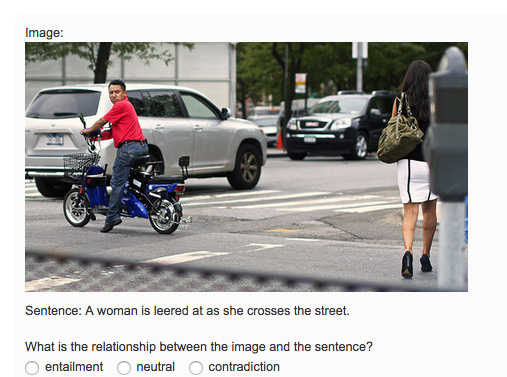}
\caption{\label{fig:ambig-leer}Ambiguous SNLI-VE instance. The lack of consensus is on whether the man is ``leering'' at the woman. While it is likely the case, this interpretation in favour of entailment is subjective, and a cautious annotator would prefer to label the instance as neutral.}
\end{figure}

\begin{figure}[!h]
\centering
\includegraphics[scale=0.3]{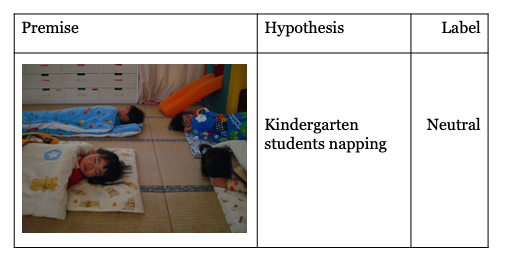}
\caption{\label{fig:ambiguous3}Ambiguous SNLI-VE instance. Some may argue that it is impossible to certify from the image that the children are kindergarten students, and label the instance as neutral. On the other hand, the furniture may be considered as typical of kindergarten, which would be sufficient evidence for entailment.}
\end{figure}

We noticed that some instances in SNLI-VE are ambiguous. We show some examples with justifications in Figures~\ref{fig:ambig-leer}, \ref{fig:ambiguous2}, and \ref{fig:ambiguous3}. In order to have a better sense of this ambiguity, three authors of this paper independently annotated 100 random examples. All three authors agreed on 54\% of the examples, exactly two authors agreed on 45\%, and there was only one example on which all three authors disagreed. We identified the following three major sources of ambiguity:
(1) mapping an emotion in the hypothesis to a facial expression in the image premise, e.g., “people enjoy talking”, “angry people”, “sad woman”. Even when the face is seen, it may be subjective to infer an emotion from a static image, (2) personal taste, e.g., “the sign is ugly”, and (3) lack of consensus on terms such as “many people” or “crowded”.

In our crowdsourced re-annotation effort, we accounted for this by removing an instance if all three annotator disagreed on the label (5.2\% for validation and 5.5\% test set). Otherwise we choose the majority label. Looking at the 18 instances where we disagreed with the label assigned by MTurk workers, we noticed that 12 were due to ambiguity in the examples, and 6 were due to workers’ errors.

\subsection{Details on Filters} \label{sec:app_filt}

In Table~\ref{tab:filters}, we provide a quantitative analysis of the effects our filters had on the dataset. The accuracies are obtained from our hand-annotated subset of 535 examples. On this subset, we first annotated every image-sentence pair as Entailment, Neutral, or Contradiction. Accuracies are obtained by comparing our own annotation with the dataset annotation. Note that we obtain higher error rates for the Entailment and Contradiction classes (9.7\% and 8.6\%) than what the authors of the original paper found~\cite{xie_visual_2019} (less than 1\%). One explanation for that could be the ambiguity that is inherent in the task. The share of bad explanations is obtained by evaluating every explanation as \emph{bad}, \emph{okay}, or \emph{great}. If the label is wrong, the explanation is automatically deemed \emph{bad}, as it will try to explain a wrong answer. 

Note that in e-SNLI, the authors have found that the human annotated explanations have an error rate of  9.6\% (19.6\% on entailment, 7.3\% on neutral, 9.4\% on contradiction), which serves as an upper bound of what could be achieved in terms of dataset cleaning.

\begin{table*}[ht!]
    \begin{center}
    \begin{tabulary}{\linewidth}{RCCCCCCCCCCC}
    \toprule
                       & \multicolumn{3}{c}{Dataset Size}  & \multicolumn{4}{c}{Share of wrong labels} & \multicolumn{4}{c}{Share of bad explanations} \\ 
    \cmidrule(r){2-4} \cmidrule(r){5-8} \cmidrule(r){9-12} 
                       & \mbox{Train Set} & \mbox{Val Set} & \mbox{Test Set} & All & E        & N        & C       & All        & E         & N         & C        \\
    \midrule
    Raw           & 529,505        & 17,554       & 17,899        & 19.3\%    & 9.7\%    & 38.6\%   & 8.6\%   & 35.7\%     & 35.2\%    & 45.1\%    & 26.3\%   \\
    FN removal         & 481,479        & 17,554       & 17,899        & 13.0\%    & 9.7\%    & 23.5\%   & 8.6\%   & 31.3\%     & 35.2\%    & 32.6\%    & 26.3\%   \\
    KW Filter          & 459,353        & 16,862       & 17,188        & 13.4\%    & 10.1\%   & 23.7\%   & 8.8\%   & 28.0\%     & 28.3\%    & 32.1\%    & 24.6\%   \\
    \mbox{Uncertainty Filter}     & 429,774        & 15,402       & 15,829        & 12.5\%    & 10.1\%   & 23.7\%   & 4.5\%   & 26.7\%     & 28.3\%    & 32.1\%    & 19.5\%   \\
    Similarity Filter  & 401,717        & 14,339       & 14,740        & 12.8\%    & 10.5\%   & 23.7\%   & 4.5\%   & 25.2\%     & 24.1\%    & 32.1\%    & 19.5\%   \\
    \bottomrule
    \end{tabulary}
  \caption{Each row describes the state of the dataset upon application of the given filter. The share of wrong labels and bad explanations is only representative of the training split. The first row describes the state of the dataset in its raw form, i.e., before any of the automatic filtering steps. The second row describes the state of the datasets upon application of the false neutral (FN) removal filter, etc.}%
  \label{tab:filters}
  \end{center}
\end{table*}

%We also find that entailment explanations are particularly frequently bad. Besides the fact that images contain more information than a single caption, another source of error is that explanations are particularly tailored to the textual premise. Indeed, Camburu et al.~\cite{camburu_e-snli_2018} had specifics check in place to make sure the explanations are based on the premise, such as the ... to use the words from the premise in the explanation.

\begin{figure}[H]
\begin{center}
\includegraphics[width=0.8\linewidth]{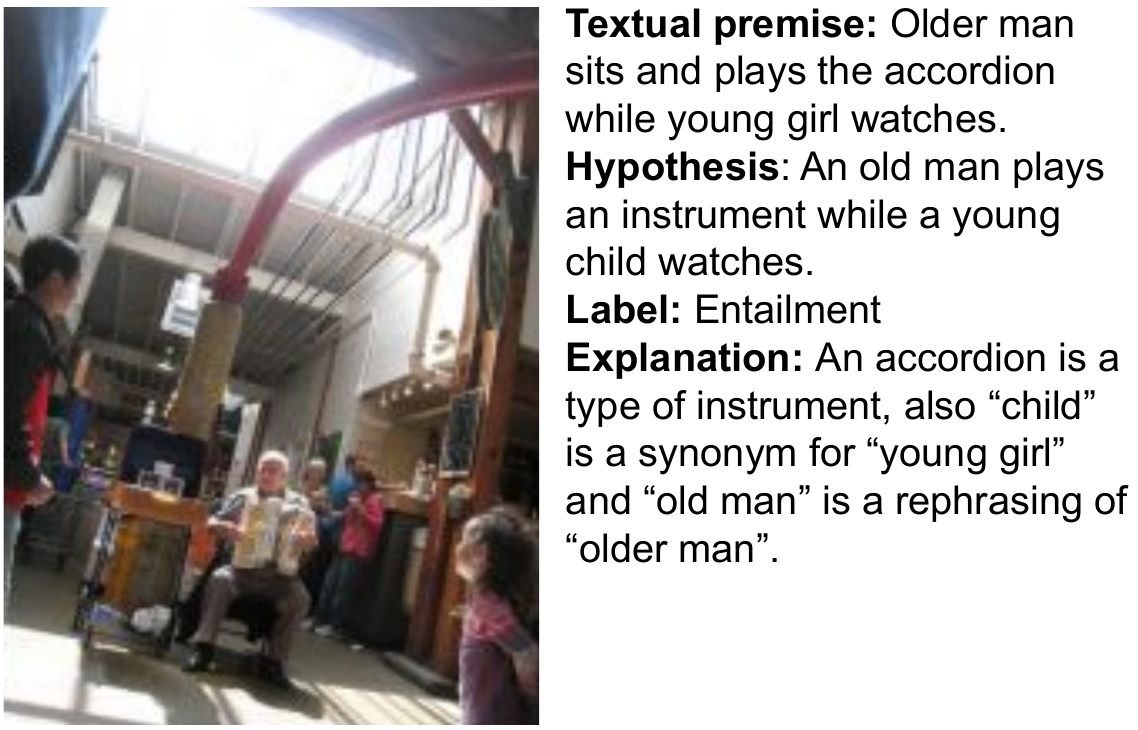}
\end{center}
   \caption{The use of the words ``synonym" and ``rephrasing" makes it clear that the explanation is overly focused on the linguistic features of the textual premise.}%  and therefore not relevant for the image
\label{fig:kw}
\end{figure}

\begin{figure}[H]
\begin{center}
\includegraphics[width=0.8\linewidth]{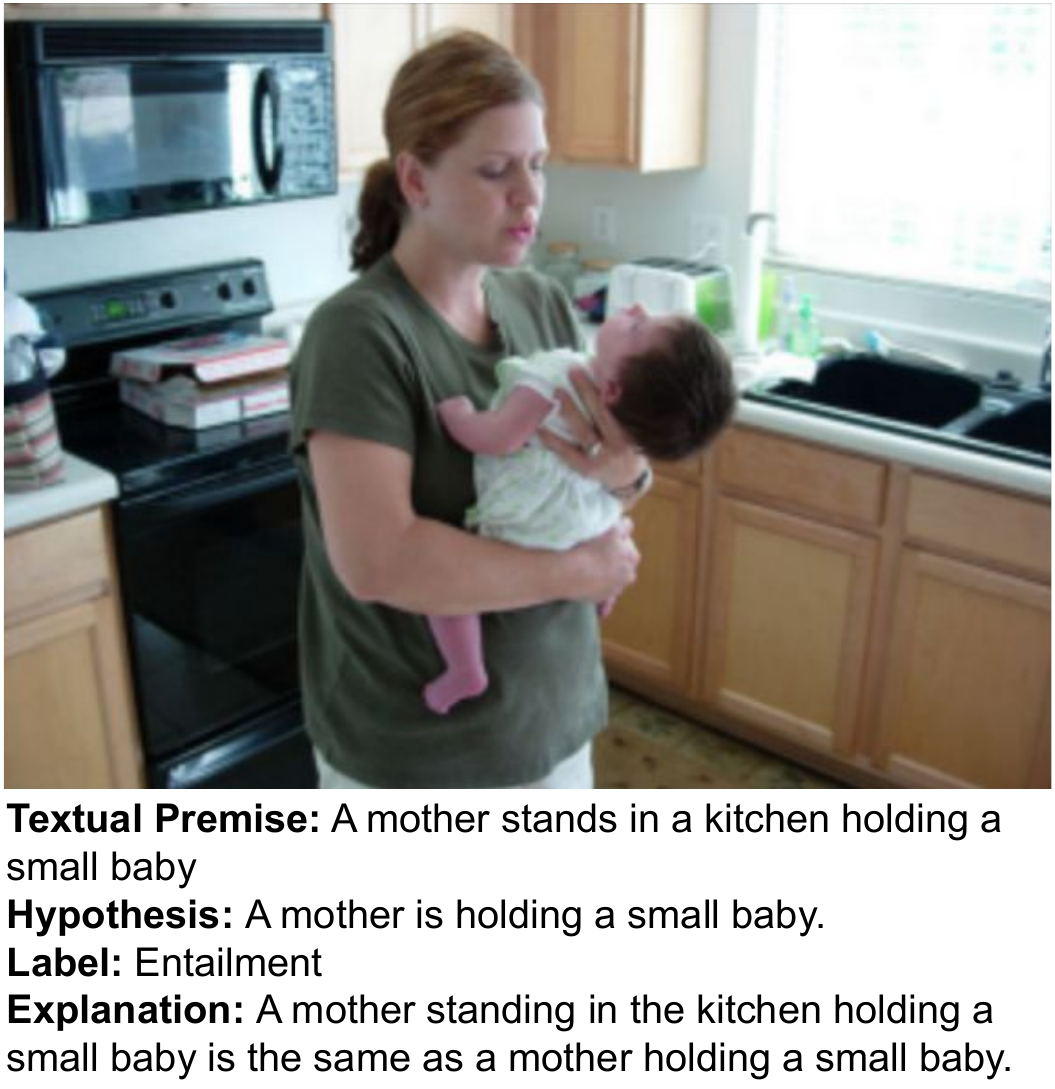}
\end{center}
   \caption{The textual premise and hypothesis are almost identical sentences, which led to a low-quality explanation.}% that do not contain useful information.}
\label{fig:sf}
\end{figure}

An illustrative example for the motivation of the false neutral detector is given in the main paper in Figure~\ref{fig:fn}. Examples for the keyword and similarity filters are given in Figures \ref{fig:kw} and \ref{fig:sf}, respectively.

%% file: appendix/datasheet.tex
The questions in this section will be answered predominantly with respect to the changes that were applied on top of (e-)SNLI, SNLI-VE, and Flickr30k. We use the datasheet form from~\citet{gebru2018datasheets}.

\subsubsection{Motivation}
\paragraph{For what purpose was the dataset created?}
The dataset was created for the purpose of extending the range of existing VL-NLE datasets with a large-scale dataset that requires fine-grained reasoning.
\vspace{-2.5ex}

\paragraph{Who created the dataset (e.g., which team, research group) and on behalf of which entity (e.g., company, institution, organization)?}
The dataset was created by researchers from the University of Oxford. It builds on existing datasets which involved other institutions (NEC Laboratories America for SNLI-VE) and universities (Stanford University for SNLI, University of Illinois at Urbana-Champaign for Flickr30k, University of Oxford for e-SNLI).

\subsubsection{Composition}
\paragraph{What do the instances that comprise the dataset represent (e.g., documents, photos, people, countries)?}
Photos (some with people) and natural language sentences.
\vspace{-2.5ex}

\paragraph{How many instances are there in total?}
In total, there are 430,796 instances.
\vspace{-2.5ex}

\paragraph{Does the dataset contain all possible instances or is it a sample (not necessarily random) of instances from a larger set? }
The dataset contains a reduced sample of the original 570k sentence pairs from SNLI \cite{bowman2015large}. It has been reduced because various filtering methods were applied to remove noise that occurred from combining e-SNLI and SNLI-VE. The filtering steps disproportionately affect the ``neutral" class. 
\vspace{-2.5ex}

\paragraph{What data does each instance consist of?}
Each instance consists of an image, a natural language hypothesis, a label that classifies the image-hypothesis pair as entailment, contradiction, or neutral, and a natural language explanation that explains why the label was given. 
\vspace{-2.5ex}

\paragraph{Is any information missing from individual instances?}
No, all instances contain the complete the information described above.
\vspace{-2.5ex}

\paragraph{Are relationships between individual instances made explicit?}
Yes. Some instances refer to the same image, which is indicated via their image ID.
\vspace{-2.5ex}

\paragraph{Are there recommended data splits?}
Yes, the train, dev, and test splits are given with the release of the dataset.
\vspace{-2.5ex}

\paragraph{Are there any errors, sources of noise, or redundancies in the dataset?}
The labels and explanations were originally annotated for textual premise-hypothesis pairs. By replacing the textual premise with an image, noise occurs. Despite our best efforts to filter out this noise, a considerable error rate remains.
\vspace{-2.5ex}

\paragraph{Is the dataset self-contained, or does it link to or otherwise rely on external resources?}
The dataset needs to be linked with Flickr30k images, which are publicly available.
\vspace{-2.5ex}

\paragraph{Does the dataset contain data that might be considered confidential (e.g., data that is protected by legal privilege or by doctor-patient confidentiality, data that includes the content of individuals’ non-public communications)? }
No.

\cleardoublepage
\subsubsection{Collection Process}

\paragraph{How was the data associated with each instance acquired?}

Hypothesises and explanations were annotated by people. SNLI-VE combined e-SNLI and Flickr30k by replacing the textual premise by an image. This was possible because the textual premises in SNLI are all captions of Flickr30k images. e-SNLI-VE was obtained by associating the explanations from SNLI with SNLI-VE. We used MTurk to reannotate the labels and explanations for the neutral class in the validation and test set. Numerous validation steps have been used to measure the effectiveness of merging, re-annotating, and filtering the dataset.
\vspace{-2.5ex}

\paragraph{What mechanisms or procedures were used to collect the data (e.g., hardware apparatus or sensor, manual human curation, software program, software API)?}
Software program and manual human curation.

%\paragraph{Who was involved in the data collection process (e.g., students, crowdworkers, contractors) and how were they compensated (e.g., how much were crowdworkers paid)?}
%The crowdworkers that reannotated the labels and explanations for the neutral class in the validation and test set were compensated \alert{TODO}\$ per hour.

\subsubsection{Preprocessing/Cleaning/Labeling}

\paragraph{Was any preprocessing/cleaning/labeling of the data done (e.g., discretization or bucketing, tokenization, part-of-speech tagging, SIFT feature extraction, removal of instances, processing of missing values)?}
Various filters were used to remove noise. We used a false neutral detector (details in Section~\ref{d:fn}), a keyword filter (details in Section~\ref{d:kw}), a similarity filter (details in Section~\ref{d:sim}), and an uncertainty filter (details in Section~\ref{d:unc}). We also reannotated all neutral examples in the validation and test set. % by reannotating their label and explanation.

\subsubsection{Distribution}
\paragraph{Will the dataset be distributed to third parties outside of the entity (e.g., company, institution, organization) on behalf of which
the dataset was created?}
The dataset is publicly released and free to access.

\subsubsection{Maintenance}
\paragraph{Who is supporting/hosting/maintaining the dataset?}
The first author of this paper.
\vspace{-2.5ex}

\paragraph{How can the owner/curator/manager of the dataset be contacted (e.g., email address)?}
The first author of this paper can be contacted via the email address given on the title page.

%% file: appendix/models.tex
This section contains further details on the models that are compared in this benchmark.

\subsection{Model Architectures}

\paragraph{PJ-X.}

The PJ-X model~\cite{park_multimodal_2018} provides multimodal explanations for VQA tasks and was originally evaluated on VQA-X. Its $M_T$ module consists of a simplified MCB network~\cite{fukui2016multimodal} that was pre-trained on VQA v2. 

% this simplified version achieves similar performance as the original MCB network.
%An element-wise multiplication of visual and language features yields a multimodal embedding, that, combined with an attention mechanism, yields the final answer.

%To generate an NLE, the combined visual and language inputs, as well as the embedded answer predicted by the model are fed to a separate visual attention operation over the visual features. 
%The weights of the explanation attention are the visual part of the multimodal explanations, as they visually \textit{point} to the relevant parts of the image.
%Similarly to the VQA part, the output of the attention operation is again combined with the embedded answer and the question embedding, and is then used as the conditioning for a two-layer LSTM~\cite{hochreiter1997long}, which decodes the explanation auto-regressively.

We implemented PJ-X in PyTorch following closely the authors' implementation in Caffe\footnote{\url{https://github.com/Seth-Park/MultimodalExplanations}}. 
To address numerical optimization problems,
%we adapted the architecture slightly.
%, which is explained in the appendix. % (Vanishing Gradident problem)
%First, 
we replaced the L2 normalization in the decoder with LayerNorm~\cite{ba2016layer}, as the original normalization zeroed gradients for earlier model parts.
Additionally, we added gradient clipping of 0.1 to prevent too large gradients.
To adapt PJ-X for multiple-choice question-answering in VCR, we follow the approach in the original VCR paper~\cite{zellers_recognition_2019}.
%All other optimization settings were kept unchanged as in the original code.

\vspace{-2ex}
\paragraph{FME.}

The model introduced by~\citet{wu_faithful_2019}, which we will refer to as FME (Faithful Multimodel Explanations), puts emphasis on producing faithful explanations. 
In particular, it aims to ensure that the explanation utilizes the same visual features that were used to produce the answer. Their code is not publicly available and we, therefore, re-implemented their base model according to the instructions in the paper. We chose the base model, as it was trained on the entire VQA-X 29.5K train split and the modifications of the other variations were difficult to re-implement from the descriptions in the paper. 
% The base model does not include their faithfulness loss nor their source identifier loss. The former did not improve the performance of our models and the latter is not feasible with the features we are using. Furthermore, it is reported to only accelerate convergence by the authors.
Our re-implementation of FME is based on a frozen modified UpDown~\cite{anderson2018bottom} VQAv2 pre-trained VL-model.
%A question-guided attention over the input image is combined with a question embedding to produce a joint multimodal representation, which is used to predict the VQA answer.

%To produce an explanation, an approach similar to the UpDown~\cite{anderson_bottom-up_2018} captioning module is used.
%First, a single answer is sampled from the normalized answer prediction.
%Its embedding is combined with the visual features and the question embedding to faithfully present the VQA focus.
%An attention LSTM is conditioned on this faithful representation as well as the last predicted explanation tokens.
%The hidden state of this attention LSTM is input to another visual attention, and a second LSTM layer auto-regressively predicts the explanation tokens.
Similarly to PJ-X, we also train FME with a gradient clipping of 0.1. % to address numerical instabilities.
To adapt FME for multiple-choice QA in VCR, we follow the approach in the original VCR paper \cite{zellers_recognition_2019}.  

\vspace{-2ex}
\paragraph{RVT.}

The Rationale-VT Transformer (RVT) model ~\cite{marasovic_natural_2020} uses varying vision algorithms to extract information from an image and then feeds this information, the ground-truth answer, and the question to a pre-trained GPT-2 language model \cite{radford2019language}, which yields an explanation. %GPT-2 is used because of its ability to condition on previous tokens, produce well-formed sentences~\cite{see2019massively} and because it harbors general knowledge~\cite{petroni2019language}. 
As they omit the question answering part, we extend their model by an answer prediction module to allow for a fair comparison and to get a sense of the overall performance. We use their overall most effective visual input\footnote{It obtained the highest visual plausibility score averaged across all datasets.}, which are the tags of the objects detected in the image. As task model $M_T$, we use BERT \cite{devlin_bert_2019}, which takes as input the object tags and the question, and predicts the answer.

\subsection{Joint or Separate Training.} \label{app:jointSep}
All the VL-NLE models $M$ in this work consist of $M_T$ and $M_E$ modules, which can either be trained jointly or separately. %Therefore, one can investigate the influence that the explanations have on a model's performance on the original task, i.e., whether the performance of $M_T$ changes when it is trained jointly with $M_E$. Some works showed that the explanations helped improve their model's performance \citep{explainyourself} while others show that, on the contrary, training jointly the explanations has a (slightly) negative effect on the performance on the original task \citep{camburu_e-snli_2018}. 
For the RVT model, training jointly would make no difference, as the explanation generation is not conditioned on a learnable representation in $M_T$ (but instead on the fixed object tags for each image). For all other models, training jointly can be advantageous, because we backpropagate the explanation loss into the task model $M_T$, but this also comes at the risk of averse effects on the optimization \cite{caruana1997multitask}. 
The authors of the PJ-X model mentioned that they tried both training approaches, but they do not specify which one worked best. \citet{wu_faithful_2019} only trained separately. It should be noted that PJ-X and FME were both solely run on VQA-X, where a much larger dataset VQA v2 exists for task $T$. They pre-train $M_T$ separately on this dataset, and it could be argued that, when training jointly, $M_T$ runs the risk of becoming worse by overfitting on the smaller dataset VQA-X. For e-SNLI-VE and VCR, no such pre-training dataset exists. In this work, we train both jointly and separately for every model.

\subsection{Reproducing Previous Results}

In this work, we reproduced three different models. The code for RVT was publicly available and we only had to add a classifier that is suited for the input type of RVT. The code of PJ-X is also publicly available, albeit in an outdated version of the Caffe framework, and therefore we translated it into Pytorch. For FME no code is available and thus we re-implemented their model (as much as possible) according to the instructions given in the paper~\cite{wu_faithful_2019}. In Table~\ref{tab:reprod} we show that the NLG metrics of our re-implementations come very close to those reported in the original papers.

For PJ-X and FME, we had to make a few minor deviations from the original implementations. To address issues with the gradients (vanishing and destabilizing) in PJ-X, we changed the L2 normalization to layer normalization~\cite{ba2016layer} in the decoder, and added gradient clipping with a threshold of 0.1. FME was re-implemented in contact with the first author of the original paper. We re-implemented their ``base'' model, which leaves out some of their model extensions. This is motivated by the fact that these extensions either did not lead to performance increases for us (their $\mathcal{L}_F$ loss) or are difficult to reproduce from the descriptions in the paper (their dataset filter $\mathcal{F}$). For the sake of standardization, we use a ResNet-101 as feature extractor for both models. We also tried a ResNet-152, but this had little effect on our results.

\begin{table*}[ht!]
    \begin{center}
    \begin{tabulary}{\linewidth}{LLCCCCC}
    \toprule
    Model                 &        & BLEU-4 & METEOR & ROUGE-L & CIDEr & SPICE \\
    \midrule
    \multirow{2}*{PJ-X~\cite{park_multimodal_2018}} & \emph{Original} & 19.8   & 18.6   & 44.0    & 73.4  & 15.4  \\
                          & \emph{Ours}   & 20.1   & 18.3   & 43.0    & 71.8  & 15.3  \\
    \multirow{2}*{FME~\cite{wu_faithful_2019}}  & \emph{Original} & 23.5   & 19.0   & 46.2    & 81.2  & 17.2  \\
                          & \emph{Ours}   & 20.8   & 19.2   & 44.8    & 77.9  & 16.7  \\
    \bottomrule
    \end{tabulary}
  \caption{A comparison (under the same settings) of automatic NLG metrics on VQA-X between our re-implementations (\emph{Ours}) of PJ-X and FME and the results reported in the papers (\emph{Original}).}%
  \label{tab:reprod}
  \end{center}
\end{table*}

\subsection{Hyperparameters} \label{app:hyp}

In total, we have four models and three datasets. For PJ-X and FME, we choose the same hyperparameters as the authors across all datasets. For PJ-X, we also experimented with larger learning rates, as we experienced convergence issues. For RVT and e-UG, we conducted grid search on three batch sizes, three learning rates, and three ways to combine the loss. We compared dynamic weight loss~\cite{liu2019end} (with two loss temperatures $T=2$ and $T=0.5$) with simply adding both losses. However, this did not affect our results enough to warrant the increase in complexity. We selected the best configuration on VQA-X and then used these settings to train on e-SNLI-VE and VCR. For BERT on VCR, we had to use a higher batch size (128), as the results would not have converged otherwise. The final hyperparameters for all four models are reported in Table \ref{tab:hparams}.

\begin{table*}[ht!] 
    \begin{center}
    \begin{tabulary}{\linewidth}{LCCCC}
    \toprule
                           & PJ-X                          & FME                           & RVT              & e-UG                          \\
    \midrule
    Batch Size             & 128                           & 128                           & 32* / 64         & 64                            \\
    Learning Rate (LR)     & \num{7e-4}                    & \num{5e-4}                    & \num{5e-5}       & \num{2e-5}                    \\
    Training Type          & JOINT*                        & JOINT*                        & SEPARATE         & JOINT                         \\
    Loss Combination       & $\mathcal{L}_T+\mathcal{L}_E$ & $\mathcal{L}_T+\mathcal{L}_E$ & N.A.             & $\mathcal{L}_T+\mathcal{L}_E$ \\
    Optimizer              & Adam                          & Adam                          & AdamW            & AdamW for BERT                \\
    LR Scheduler           & -                             & Step decay                 & Linear w/ warmup & Linear w/ warmup              \\
    Tokenization           & Word                          & Word                          & WordPiece        & WordPiece                     \\
    Max Question Length    & 23                            & 23                            & 19               & 19                            \\
    Max Answer Length      & 23                            & 40                            & 23               & 23                            \\
    Max Explanation Length & 40                            & 40                            & 51               & 51                            \\
    Decoding               & Greedy                        & Greedy                        & Greedy           & Greedy                        \\
    \bottomrule
    \end{tabulary}
  \caption{Hyperarameters used for the different models across all datasets. $\mathcal{L}_T$ and $\mathcal{L}_E$ are the task loss and explanation loss, respectively. For RVT, the task batch size for VCR is 128, as 32 did not lead to convergence.  For PJ-X and FME, we trained $M_T$ and $M_E$ separately on VQA-X.}%
  \label{tab:hparams}
  \end{center}
\end{table*}

An additional overview of the differences between the models is given in Table~\ref{tab:modDiff}.

\begin{table*}[ht!]
    \begin{center}
    \begin{tabulary}{\linewidth}{LLLLL}
    \toprule
    Model $M$ & Vision Backbone & VL Model $M_T$ & Explanation Model $M_E$ & $M_E$ Input     \\
    \midrule
    PJ-X & ResNet-101 & MCB & LSTM (a) & image features, question, answer            \\
    FaiMu & ResNet-101   & UpDown    & LSTM (b)     & image features, question, answer          \\
    RVT   & \mbox{Faster R-CNN} & BERT   & GPT-2   & object tags, question, answer             \\
    e-UG  & \mbox{Faster R-CNN} & UNITER   & GPT-2  & contextualized embeddings of image-question pair, question, answer \\
    \bottomrule
    \end{tabulary}
  \caption{Summary of the model differences.}%
  \label{tab:modDiff}
  \end{center}
\end{table*}

%\subsection{Teacher forcing of answers}

%For all our approaches the explanation generator $M_E$ is conditioned on the answer of task $T$. In the literature, there are three approaches to do this. In the first one (1), we feed the predicted answer $a=M_T(V,L)$ to the explanation generator $M_E$. $M_E$ then predicts $e$, based off the final representation of  $M_T$ and the answer $a$. An issue with that approach is that, in case $a$ is predicted incorrectly, we train the model to generate an explanation for a different label than the model is conditioned on. In the second option (2), we only generate the explanations for examples where $a$ was correct. That way, explanations are always conditioned on the right label, but we waste the training signal from all the explanations where the label was incorrect, which can be substantial. Finally, in the third option (3), we always condition $M_E$ on the ground-truth answer during training. This way, we make use of all the training data and the explanations are always conditioned on the right answer. This is conceptually similar to teacher forcing in language modelling and while this is not perfect (exposure bias) it is still accepted as the most commonly used method. For fair comparison, we opt for method 3 and use it across all tasks.  PJ-X used approach (1) and FaiMu used approach (2). We have confirmed that the model's performance is not significantly impacted by changing the training regime. 

\subsection{Adaptations for VCR}

To accommodate for the multiple-choice nature of task~$T$, we adapt the architectures accordingly. For UNITER, we follow the original paper and formulate multiple-choice as a binary classification of question-image-answer tuples as True or False. The final answer is determined through a softmax of the four True scores. For PJ-X and FME, we follow the approach in the original VCR paper and obtain the logit for response $j$ via the dot product of the final representation of the model and the final hidden state of the LSTM encoding of the response $r^j$ \cite{zellers_recognition_2019}. For RVT, we use \textsc{BertForMultipleChoice} from the transformers library~\cite{wolf-etal-2020-transformers}.

%% file: appendix/eval.tex
%\subsection{Anchor Effects}

%When using human evaluation for multiple models, we run the risk of introducing anchoring effects that can negatively affect the re-usability of the e-ViL Framework. Consider this scenario: we have four models, three of which are quite bad and one is \emph{okay}, but not great. If an annotator sees a lot of bad explanations, he could start lowering his standards and be overly generous to the model that is \emph{okay}, but not great. If that \emph{okay} model would be evaluated on its own, it would perhaps obtain lower scores. Thus, some future model might be equally good than the \emph{okay} one, but receive lower scores, making the benchmark less re-usable. For this reason, why do not mix any models in the human evaluation questionnaires and evaluate each model separately.

%An additional step that we take to ensure the same anchoring effect
%Further, we attempt to apply the same anchoring effect to all evaluations, by always asking an evaluation of the GT and predicted explanation, without workers knowing which is which.

\begin{figure*}[ht]
\begin{center}
\includegraphics[scale=0.6]{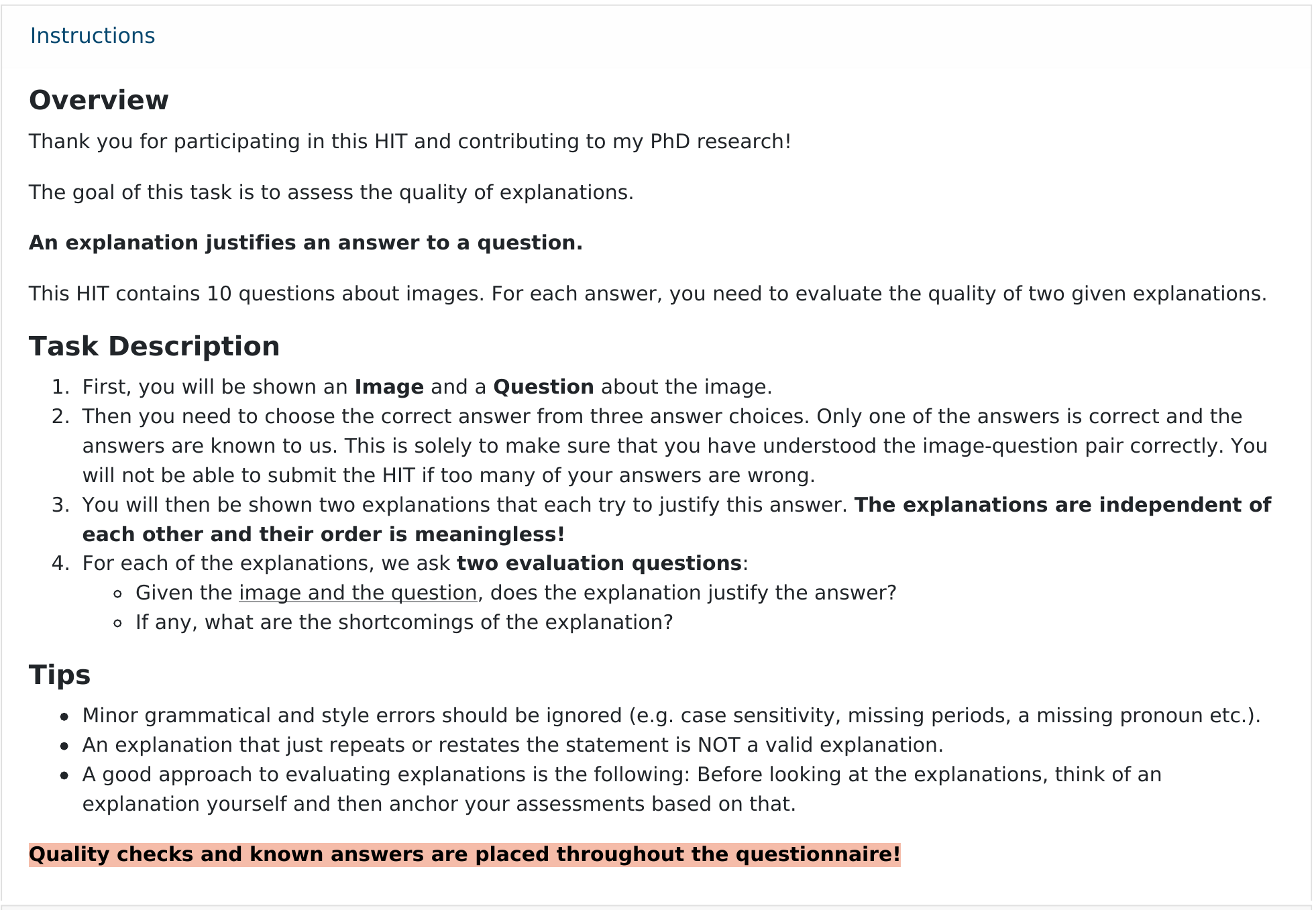}
\end{center}
   \caption{A snapshot of the instructions that were provided to the annotators that evaluated the explanations.}%
\label{fig:evil_inst}
\vspace{-2ex}
\end{figure*}

An example of the instructions that were shown to the MTurk annotators can be seen in Figure~\ref{fig:evil_inst}. The interface through which the annotators evaluated the explanations is displayed in Figure~\ref{fig:evil_inter}. %We will release the fully reproducible MTurk HTML files upon acceptance. 
The cost to evaluate \emph{one} model on \emph{one} dataset is 108-117\$.

\begin{figure}[ht!]
\begin{center}
\includegraphics[width=1\linewidth]{app_figures/vqax_mturk_example.png}
\end{center}
   \caption{A snapshot of the interface through which annotators evaluated the explanations.}%
\label{fig:evil_inter}
\vspace{-2ex}
\end{figure}

 % Should we include the argumentation for why we did it differently than Marasovic et al.? Or more generally how we picked best practices from existing evaluation frameworks? ... Marasovic et al. evaluates their explanations by asking three questions: (1) Is the explanation relevant given the question? (2) Is the explanation relevant given the image and question? (3) Does the explanation refer to xx not present in the image? We think a better way of assessing explanations is to first 
 
 %We provide the MTurk evaluation scripts @ xx. At an hourly rate of xx\$, and having three annotators per datapoint, this amounts to xx\$ to get an e-ViL score per model. This can be reduced by not asking the workers to list the shortcomings of the explanation. 
%Upon acceptance we will also release all our human-evaluated explanations to enable potential future research into automated metrics. 

%% file: appendix/results.tex
In this section, we present a benchmark evaluation with automatic NLG metrics (\ref{app:sec:autoNLG}), extended results on e-SNLI-VE performance (\ref{sec:esnlive_dets}) and different ways to compute the e-ViL $S_E$ score (\ref{sec:evil_alts}).

\vspace{-2ex}
\subsection{Automatic NLG Metrics} \label{app:sec:autoNLG}

We report the automatic NLG scores in Table \ref{tab:autoNLG}. Those are computed for all the explanations from the test sets where the predicted answer was correct. A quick observation is that the human evaluation results are not always reflected by the automatic metrics. For example, on the VCR dataset, FME, and not e-UG, obtains the highest $S_E$ score when using automatic NLG metrics. Some tendencies are reflected nonetheless, such as the fact that e-UG is the best model overall and that e-UG consistently outperforms RVT (albeit by a small margin). 

\vspace{-2ex}
\paragraph{Question-only GPT-2.}
In order to verify our intuition that the object labels used by RVT provide very little information about the image, we trained GPT-2 that only conditions on the question and answer, ignoring the image (called \emph{GPT-2 only} in Table \ref{tab:autoNLG}). Without having any image input, this model closely shadows the performance of RVT on most metrics. RVT is still slightly better in most cases, indicating that the object labels do provide some minor improvement. This suggests that RVT is not able to use visual information effectively and learns the explanations mostly based off spurious correlations and not based on the image.

\begin{table*}[htbp]
    \begin{center}
    \begin{tabulary}{\linewidth}{LCCCCCCCCCCCC}
    \toprule
                       & \multicolumn{3}{c}{e-ViL Scores (auto)} & \multicolumn{8}{c}{$n$-gram Scores} & Learned Sc\\
    \cmidrule(r){2-4} \cmidrule(r){5-12} \cmidrule(r){13-13}
    \emph{VQA-X}       & $S_O$ & $S_T$ & $S_E$ & B1 & B2 & B3 & B4 & R-L & MET. & CIDEr & SPICE & BERTScore \\
    \midrule
    PJ-X~\cite{park_multimodal_2018} & 32.1  & 76.4  & 42.1  & \textbf{57.4}   & 42.4   & 30.9   & 22.7   & \textbf{46.0}    & 19.7   & 82.7  & 17.1  & 84.6      \\
    FME~\cite{wu_faithful_2019} & 33.0  & 75.5  & 43.7  & 59.1   & \textbf{43.4}   & \textbf{31.7}   & 23.1   & 47.1    & 20.4   & \textbf{87.0}  & 18.4  & 85.2      \\
    RVT~\cite{marasovic_natural_2020} & 26.8  & 68.6  & 39.1  & 51.9   & 37.0   & 25.6   & 17.4   & 42.1    & 19.2   & 52.5  & 15.8  & 85.7      \\
    GPT-2 only & N.A.  & N.A.  & 37.8  & 51.0   & 36.4   & 25.3   & 17.3   & 41.9    & 18.6   & 49.9  & 14.9  & 85.3      \\
    e-UG       & \textbf{36.5}  & \textbf{80.5}  & \textbf{45.4}  & 57.3   & 42.7   & 31.4   & \textbf{23.2}   & 45.7    & \textbf{22.1}   & 74.1  & \textbf{20.1}  & \textbf{87.0}    \\
    \midrule
    \emph{VCR} &  &  & &  & &  &  &  &  &  &  & \\
    \midrule
    PJ-X~\cite{park_multimodal_2018} & 7.2   & 39.0  & 18.4  & 21.8   & 11.0   & 5.9    & 3.4    & 20.5    & 16.4   & 19.0  & 4.5   & 78.4      \\
    FME~\cite{wu_faithful_2019} & 17.0  & 48.9  & \textbf{34.8}  & \textbf{23.0}   & \textbf{12.5}   & \textbf{7.2}    & \textbf{4.4}    & \textbf{22.7}    & \textbf{17.3}   & 27.7  & \textbf{24.2}  & \textbf{79.4}      \\
    RVT~\cite{marasovic_natural_2020} & 15.5  & 59.0  & 26.3  & 18.0   & 10.2   & 6.0    & 3.8    & 21.9    & 11.2   & 30.1  & 11.7  & 78.9      \\
    GPT-2 only & N.A   & N.A   & 26.3  & 18.0   & 10.2   & 6.0    & 3.8    & 22.0    & 11.2   & 30.6  & 11.6  & 78.9      \\
    e-UG       & \textbf{19.3}  & \textbf{69.8}  & 27.6  & 20.7   & 11.6   & 6.9    & 4.3    & 22.5    & 11.8   & \textbf{32.7}  & 12.6  & 79.0    \\
    \midrule
    \emph{e-SNLI-VE} &  &  &  &  &  &  &  &  &  &  &  &  \\
    \midrule
    PJ-X~\cite{park_multimodal_2018} & 26.5  & 69.2  & 38.4  & 29.4   & 18.0   & 11.3   & 7.3    & \textbf{28.6}    & 14.7   & 72.5  & 24.3  & 79.1      \\
    FME~\cite{wu_faithful_2019} & 29.9  & 73.7  & 40.6  & \textbf{30.6}   & 19.2   & 12.4   & 8.2    & 29.9    & 15.6   & 83.6  & 26.8  & 79.7      \\
    RVT~\cite{marasovic_natural_2020} & 31.7  & 72.0  & 44.0  & 29.9   & 19.8   & 13.6   & \textbf{9.6}    & 27.3    & 18.8   & 81.7  & 32.5  & 81.1      \\
    GPT-2 only       & N.A.  & N.A.  & 43.6  & 29.8   & 19.7   & 13.5   & 9.5    & 27.0    & 18.7   & 80.4  & 32.1  & 81.1      \\
    e-UG             & \textbf{36.0}  & \textbf{79.5}  & \textbf{45.3}  & 30.1   & \textbf{19.9}   & \textbf{13.7}   & \textbf{9.6}    & 27.8    & \textbf{19.6}   & \textbf{85.9}  & \textbf{34.5}  & \textbf{81.7}  \\
    \bottomrule
    \end{tabulary}
  \caption{Automatic NLG metrics for all model-dataset pairs. The $S_E$ based on automatic NLG metrics is the harmonic mean that was used to select the best model during validation. B1 to B4 stand for BLEU-1 to BLEU-4, R-L for ROUGE-L, and MET for METEOR.}%
  \label{tab:autoNLG}
  \end{center}
\end{table*}

\subsection{Detailed Results for e-SNLI-VE} \label{sec:esnlive_dets}

Here, we provide more detailed results on our newly released e-SNLI-VE dataset. We break down the task accuracy and explanation scores by the three different classes (see Table~\ref{tab:esnlive_det}). For all models, we observe significantly lower accuracies and explanation scores for the neutral class. There are two potential explanations for this. First, the neutral class can be harder to identify than the other classes. In image-hypothesis pairs, entailment and contradiction examples can sometimes be reduced to more straightforward yes/no classifications of image descriptions. For the neutral class, there always needs to be some reasoning involved to decide whether the image does (not) contain enough evidence to neither indicate entailment nor contradiction. A second reason is that, despite our best efforts to clean the dataset, the neutral class is still more noisy and less represented in the training data.

\begin{table*}[htbp]
    \begin{center}
    \begin{tabulary}{\linewidth}{LCCCCCCCCC}
    \toprule
         & \multicolumn{3}{c}{Entailment}                & \multicolumn{3}{c}{Neutral}                   & \multicolumn{3}{c}{Contradiction}             \\
    \cmidrule(r){2-4} \cmidrule(r){5-7} \cmidrule(r){8-10}
         & Acc.      & MET.       & BERTS.     & Acc.      & MET.        & BERTS.     & Acc.      & MET.        & BERTS.     \\
    \midrule
    PJ-X & 74.4          & 14.0          & 79.2          & 61.5          & 12.4          & 77.4          & 72.8          & 15.9          & 79.3          \\
    FME  & 77.3          & 15.1          & 79.8          & 67.3          & 13.5          & 77.9          & 77.2          & 16.3          & 79.8          \\
    RVT  & 74.6          & 17.9          & 81.3          & 63.3          & \textbf{19.0} & 80.7          & 79.4          & 19.4          & 81.4          \\
    e-UG & \textbf{80.3} & \textbf{19.6} & \textbf{81.6} & \textbf{71.7} & 18.5          & \textbf{80.9} & \textbf{87.5} & \textbf{20.9} & \textbf{82.6} \\
    \bottomrule
    \end{tabulary}
  \caption{Class-wise results on e-SNLI-VE for the different models. NLG metrics are only shown for METEOR and BERTScore, as those correlate most with human judgement.}%
  \label{tab:esnlive_det}
  \end{center}
\end{table*}

\subsection{Statistical Analysis of the $S_E$ Score} \label{sec:stat_analysis}

To ensure high quality of our results, we had a number of in-browser checks that prevented the annotators from submitting the questionnaire when their evaluations seemed of poor quality. Checks include making sure that they cannot simultaneously say that an explanation is insufficient (they select the \emph{No} or \emph{Weak No} option described in Section \ref{sec:evil}) and has no shortcomings, or that it is optimal (they select \emph{Yes} option), but has shortcomings. We also experimented with further post-hoc cleaning measures (such as verifying that they evaluated the ground-truth favorably or did not always choose similar answers), but they had a negligible impact and thus were disregarded.

Our MTurk sample consists of 19,194 evaluations, half of which are for ground-truth explanations, and the other half for model generated explanations. We obtain evaluations for 264 to 299 unique question-image pairs for every model-dataset combination, leaving us with explanations missing for only 3.3\% of questions. There are 82.1 evaluations per annotator on average ($SD=170.1$), ranging from 16 to 1,244 with a median of 34. After pooling annotations of the same explanation, 6,494 annotations remain (887 to 897 for the evaluations generated by each model).

\begin{figure*}[ht]
    \centering
    \includegraphics[width=0.65\textwidth]{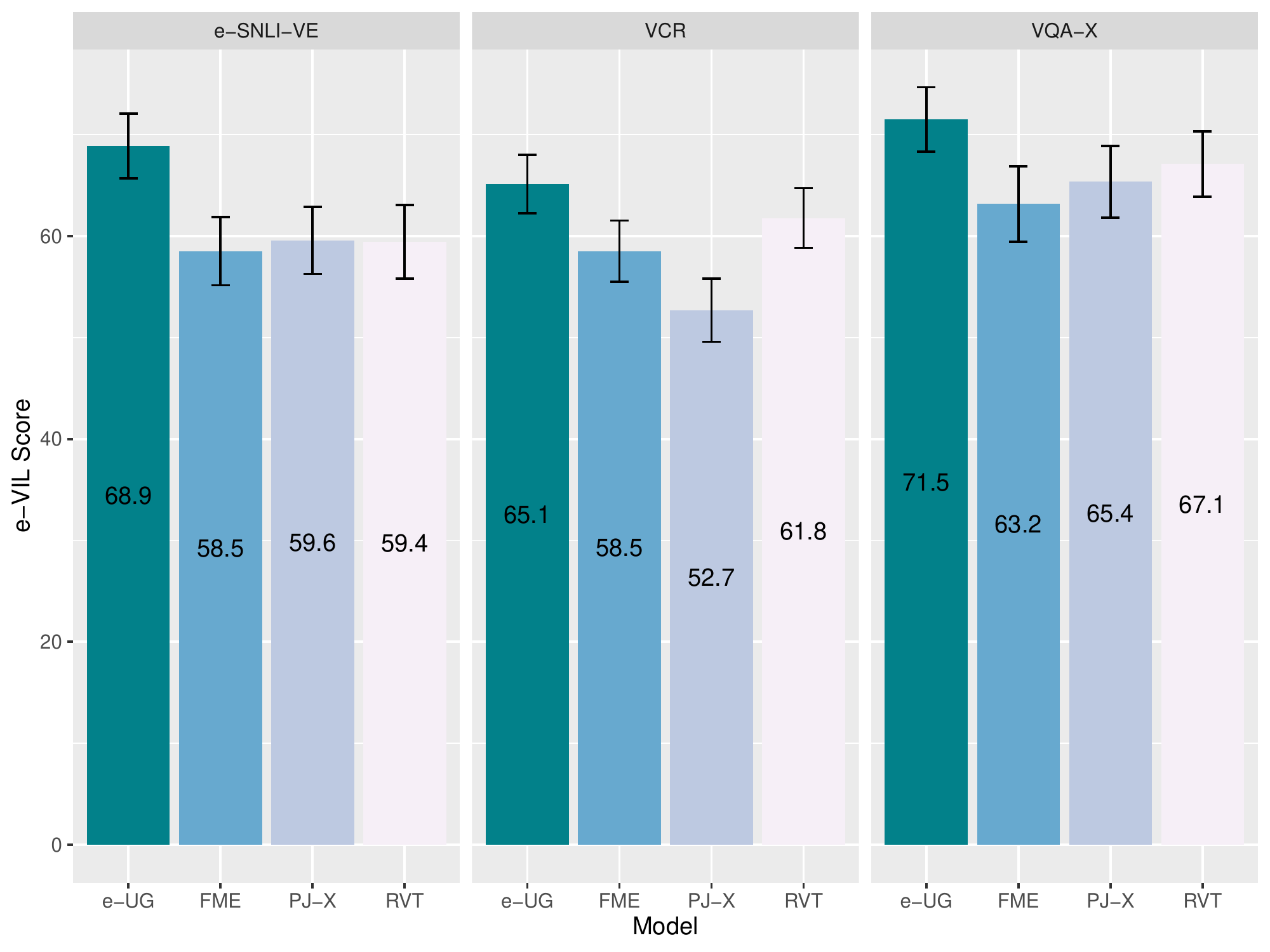}
    \caption{Human evaluation framework: e-ViL scores $S_E$. This plot shows the main e-ViL scores (based on numerical average) for the different model-dataset pairs. Error bars show $\pm 2 \text{SD} / \sqrt{n}$ for each group.}%
    \label{fig:appendix_he_evil_bar}
\end{figure*}

In Figure~\ref{fig:appendix_he_evil_bar}, we add standard errors to the numerical $S_E$ scores given in Table~\ref{tab:he_score}. This figure confirms that e-UG uniformly outperforms the other models.

To further investigate the robustness of the e-ViL benchmark, we do a statistical analysis of our $S_E$ scores by using a Linear Mixed Model (LMM) that predicts $S_E$ from the model-dataset pairs, with model as fixed factor and dataset as random effect. LMM predicts the evaluations with the Likelihood-Ratio-Test of the fixed effect being significant, with $\chi^2(3)=37.462, p<0.001$. To gain better insight, we performed post-hoc pairwise contrasts, which indicate that e-UG significantly outperforms the remaining models, with $p<0.001$. Further, RVT outperforms PJ-X significantly, with $p=0.007$. The significance level was adjusted for a family-wise type I error rate of $\alpha=0.05$ using Bonferroni-Holm adjustments.

\begin{figure*}[ht]
    \centering
    \includegraphics[width=0.65\textwidth]{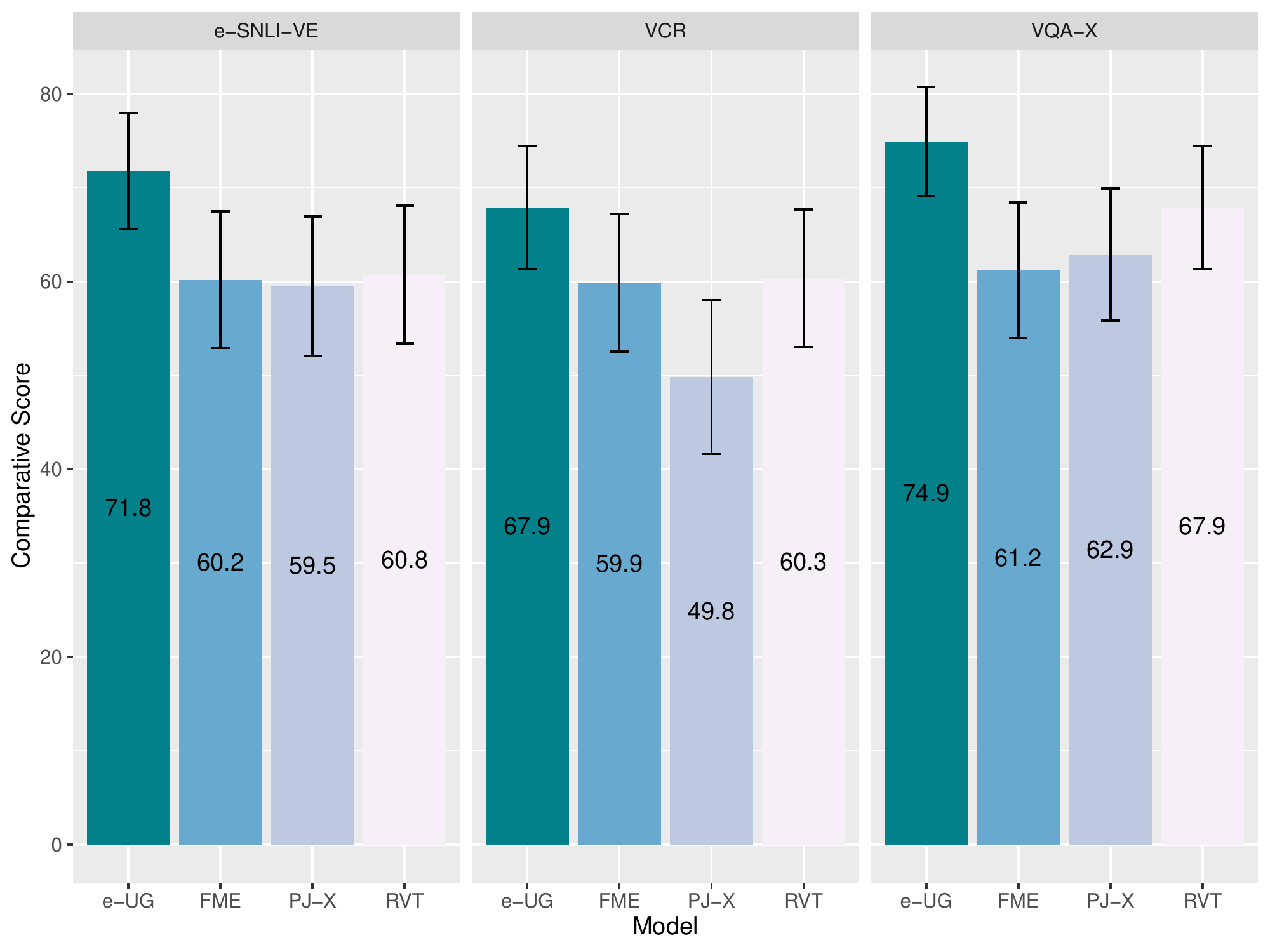}
    \caption{Human evaluation framework: Comparative scores. This figure displays the comparative scores (with respect to the ground-truth) of the explanations for the different model-dataset pairs. Error bars show $\pm 2 \text{SD} / \sqrt{n}$ for each group.}
    \label{fig:appendix_he_compare_gt}
\end{figure*}

\subsection{Alternative $S_E$ Scores} \label{sec:evil_alts}

The nature of our human evaluation questionnaire allows for multiple ways to compute the e-ViL score $S_E$ of the generated explanations. The key differences between the scoring methods are on how to pool the up-to-three evaluations we have for each explanation, and how to compute the overall numerical value. In the main paper, we compute $S_E$ by mapping the four evaluation choices to numerical values, then taking the average for every explanation in the sample and then the sample average to get our $S_E$ score. Below, we propose two alternative ways to compute $S_E$. While they lead to different values, the performance differences between our models remain relatively similar.

\subsubsection{Median Pooling}

\begin{figure*}[ht]
    \centering
    \includegraphics[width=\textwidth]{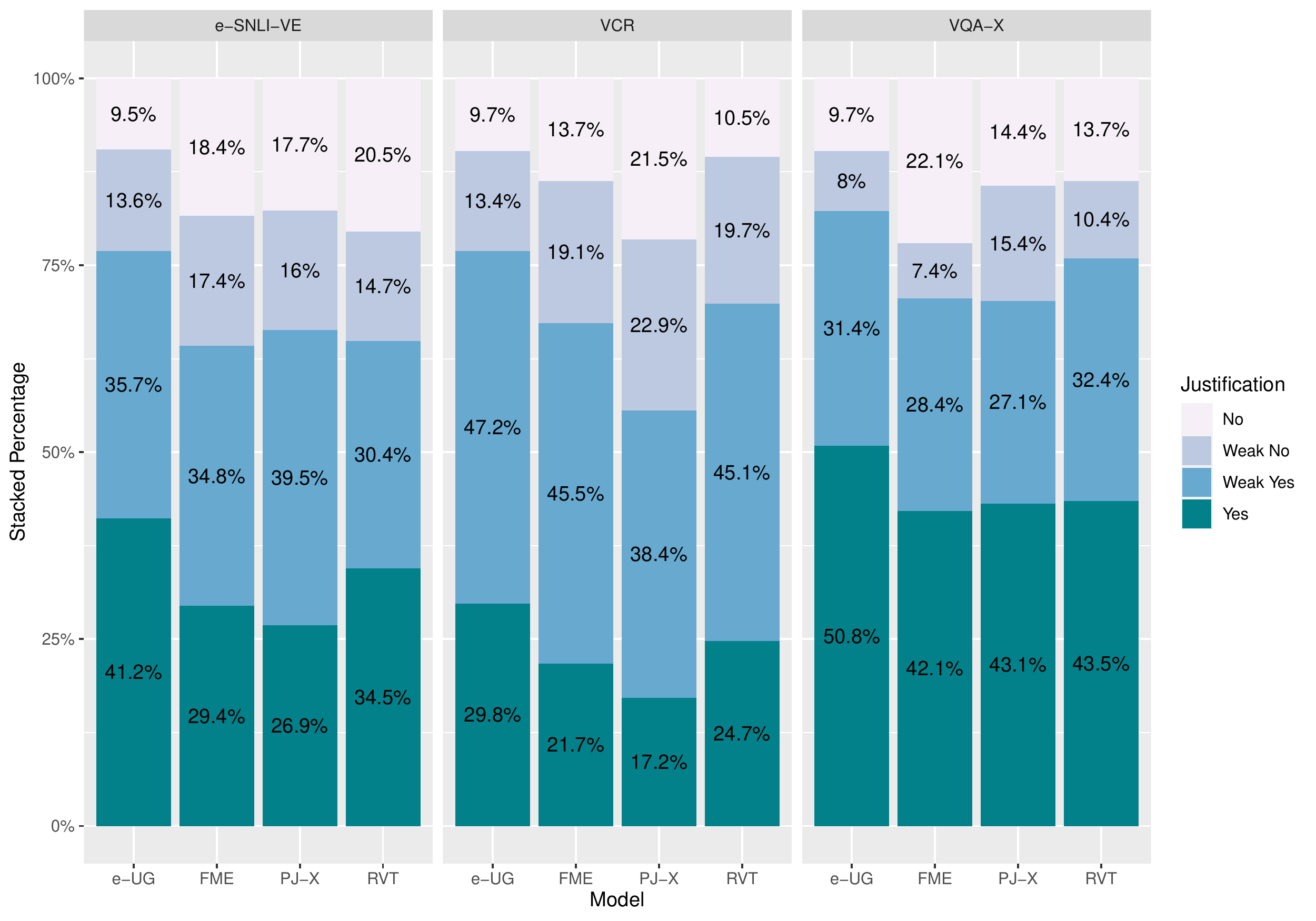}
    \caption{Human evaluation framework: Ordinal representation of the evaluations. Median responses for each question-image pair given by participants to the evaluation question question ``Given the image and the question/hypothesis, does the explanation justify the answer?".}%
    \label{fig:appendix_pooled_barplot}
\end{figure*}

In median pooling, we obtain the score for each explanation by taking the median of its up-to-three ordinal evaluations (as opposed to taking a numerical average). We always interpolate with rounding off, meaning that the median of (\emph{Yes}, \emph{Weak Yes}) $\mapsto$ \emph{Weak Yes} and (\emph{Yes}, \emph{No}) $\mapsto$ \emph{Weak No}. This allows us to plot the distribution of \emph{No}, \emph{Weak No}, \emph{Weak Yes}, and \emph{Yes} for every model-dataset pair, as displayed in Figure~\ref{fig:appendix_pooled_barplot}.

We observe that e-UG performs better across all datasets, with RVT following in second place for the VCR and VQA-X datasets. The differences between the PJ-X, FME and RVT are relatively small.

We analyse our results using a Cumulative Link Mixed Model (CLMM) with a logit link and flexible thresholding. We predict annotator responses using the dataset as random effect and the VL-NLE model as fixed effect. We find that the model significantly influences ratings, as suggested by the Likelihood-Ratio-Test, $\chi^2(2)=42.4, p < 0.001$, when comparing the full model to a nested statistical model 
%(predictors of the smaller model are a subset of the predictors of the larger model) 
that is merely based on the dataset as predictor. The model predictor is dummy-coded with e-UG as reference class %(one-hot encoding with reference class being represented with the 0 vector)
, which enables us to interpret the model's coefficients in the statistical test
%(model weights and bias with assumptions about their distribution) 
as pairwise contrasts 
%(linear system of equations that can be tested for statistical significance given the distributional assumptions of underlying coefficients) 
of all other models towards e-UG. All coefficients have $p$-values $p<0.001$, indicating the e-UG significantly outperforms all other models.

\subsubsection{Comparative $S_E$ Score}

We also designed a comparative score, for which we do not map our questionnaire evaluation options (\emph{No}, \emph{Weak No}, \emph{Weak Yes}, and \emph{Yes}) to numerical values, but instead compare them to the evaluation of the ground-truth. For every image-question pair, the annotator has to evaluate both the ground-truth and the generated explanation, without knowing which is which. This enables us to see, for every generated explanation, if it was deemed equally good, better, or worse than the ground-truth. This mimics the approach in \citet{park_multimodal_2018} and \citet{wu_faithful_2019}, where annotators were explicitly asked if the generated explanation was worse, equally good, or better than the ground-truth. An advantage of this method is that we can seamlessly incorporate the criticalness of each annotator. The disadvantage is that we do not get \emph{absolute} measurements of the quality of the explanations.

The generated explanation gets the score 1 if it is as good or better than the ground-truth, and otherwise 0. We pool the comparative score via median pooling with rounding off. 

Figure \ref{fig:appendix_he_compare_gt} displays the comparative score. We can observe that e-UG scores are strongest across all datasets, while the other three models are performing similarly, except on the VCR dataset, where PJ-X performs worse than the other models.

For our statistical analysis, we fit a generalized linear mixed model (GLMM) on the full unpooled annotation set predicting the whether an explanation was rated positively (compared to the ground-truth) using the dataset and annotator as random effects and the VL-NLE model as fixed effect. We utilise a logit link. The model parameter significantly predicts the evaluations, with $\chi^2(3)=67.366, p<0.001$. Post-hoc tests (Tukey contrasts with Bonferroni-Holm adjusted significance) show that the e-UG outperforms all other models, with $p<0.001$, and that RVT outperforming PJ-X at $p=0.011$. All other pairwise comparisons were not significant. Extending the model to include ground-truth explanations as a model category also demonstrates that all model-generated explanations were evaluated significantly worse than the ground-truth explanations. We conclude that the e-UG outperforms all other models, whereas performance differences between them are rather small, replicating our findings from the alternative analyses.

%\begin{table*}[htbp]
%    \begin{center}
%    \begin{tabulary}{\linewidth}{LCCCC}
%    \toprule
%         & Average       & VQA-X         & e-SNLI-VE     & VCR           \\
%    \midrule
%    PJ-X & 57.4          & 62.9          & 59.5          & 49.8          \\
%    FME  & 60.4          & 61.2          & 60.2          & 59.9          \\
%    RVT  & 63.0          & 67.9          & 60.8          & 60.3          \\
%    e-UG & \textbf{71.5} & \textbf{74.9} & \textbf{71.8} & \textbf{67.9} \\
%    \bottomrule
%    \end{tabulary}
%  \caption{}%
%  \label{tab:se_v2}
%  \end{center}
%\end{table*}